\documentclass[10pt,twocolumn,letterpaper]{article}

\usepackage[pagenumbers]{wacv} 

\usepackage{graphicx}
\usepackage{booktabs}
\usepackage{textcomp}
\usepackage{verbatim}

\usepackage{blindtext}
\usepackage{CJKutf8}
\usepackage{setspace}
\usepackage{algorithm}
\usepackage{algpseudocode}

\usepackage{subcaption}
\usepackage[utf8]{inputenc} 
\usepackage[T1]{fontenc} 
\usepackage{url} 
\usepackage{amsfonts} 
\usepackage{nicefrac} 
\usepackage{microtype} 
\usepackage{array}
\usepackage{amssymb}
\usepackage{mathtools}
\usepackage{amsthm}
\usepackage{multirow}
\usepackage{colortbl}
\usepackage{enumitem}
\usepackage{tocloft}
\usepackage{tabularx}
\usepackage{tcolorbox}
\usepackage{soul}
\usepackage{wrapfig}
\usepackage{siunitx}
\usepackage{pifont}
\usepackage{makecell}
\usepackage{upgreek}
\usepackage{bm}
\usepackage{fontawesome5} 

\usepackage[dvipsnames,table]{xcolor}
\colorlet{MyCol}{Tan!10}

\newcommand{\colrow}{\rowcolor{MyCol}}

\newcommand{\venue}[1]{\textcolor{darkgray}{\scriptsize{[#1]}}}

\newcommand{\DD}{\mathcal{D}}
\newcommand{\XX}{\mathbf{X}}

\newcommand{\bAA}{\mathbf{A}}
\newcommand{\idmx}{\bAA_{\text{id}}}
\newcommand{\attmx}{\bAA_{\text{att}}}
\newcommand{\QQ}{\mathbf{Q}}

\newcommand{\zz}{\mathbf{z}}
\newcommand{\ZZ}{\mathbf{Z}}

\newcommand{\pp}{\mathbf{p}}

\newcommand{\vv}{\mathbf{v}}

\newcommand{\PP}{\mathcal{P}}
\newcommand{\bA}{\mathcal{A}}

\newcommand{\qq}{\mathbf{q}}
\newcommand{\venc}{\mathcal{F}_v}
\newcommand{\tenc}{\mathcal{F}_t}
\newcommand{\tokn}{\mathcal{T}}
\newcommand{\Real}{\mathbb{R}}

\newcommand{\method}{{\texttt{PRAISE}}}

\newcommand{\PPi}{\bm{\Uppi}}

\newcommand{\mypar}[1]{
\vspace{3pt}
\noindent\textbf{#1}~}

\newcommand{\phz}{\phantom{0}}

\usepackage{amsthm}
\newtheorem{proposition}{Proposition}[section]
\definecolor{wacvblue}{rgb}{0.21,0.49,0.74}
\usepackage[pagebackref,breaklinks,colorlinks,allcolors=wacvblue]{hyperref}

\def\wacvPaperID{1156}
\def\confName{WACV}
\def\confYear{2027}

\title{
In \method{} of Attributes: Grounding Retrieval Geometry with Query-Conditioned Subspaces for Person Re-Identification}

\author{%
Rajarshi Bhattacharya$^{1}$\thanks{Corresponding author:\\ \texttt{rajarshi.bhattacharya.1@ens.etsmtl.ca}.}
\quad
Aryan Shukla$^{1}$
\quad
Shakeeb Murtaza$^{1}$\\
Christian Desrosiers$^{1}$
\quad
Jose Dolz$^{1}$
\quad
Maguelonne Heritier$^{2}$
\quad
Eric Granger$^{1}$\\[0.5em]
$^{1}$LIVIA, École de technologie supérieure\\
$^{2}$Genetec Inc.\\
\href{https://github.com/rB080/praise.git}{\faGithub Code}
}

\begin{document}
\maketitle

\begin{abstract}
Person re-identification (ReID) ranks an image gallery using either an image query or a textual query. In image-to-image (I2I) ReID, viewpoint, illumination, occlusion, and background changes can make distractors appear more similar than the true match. In text-to-image (T2I) ReID, incomplete or ambiguous descriptions may omit key visual evidence.
We study this problem for vision-language model (VLM) based ReID, where attributes such as clothing color and carried objects provide useful fine-grained matching cues. However, directly using attribute-prompt similarities as retrieval scores is unreliable because templated prompts can be biased, correlated, and poorly calibrated. We therefore propose \textbf{PR}ompt-guided \textbf{A}ttribute-based \textbf{I}nference-time \textbf{S}imilarity \textbf{E}nhancement (\texttt{PRAISE}), a query-side test-time adaptation framework for VLM-based ReID. Instead of treating prompts as scores, \texttt{PRAISE} uses a transferable attribute-prompt bank to construct an \emph{Attribute-Conditioned Subspace} (ACS): a query-dependent low-rank subspace induced by the most compatible attribute prompts. ACS provides a controlled attribute-conditioned geometry for reshaping query--gallery affinities and guiding lightweight inference-time updates without identity or attribute labels.
Experiments across multiple I2I- and T2I-ReID benchmarks show that \texttt{PRAISE} consistently improves VLM retrieval backbones and outperforms both generic and ReID-specific TTA baselines.
\end{abstract}

\begin{figure}[!h]
    \centering
    \includegraphics[width=0.77\linewidth]{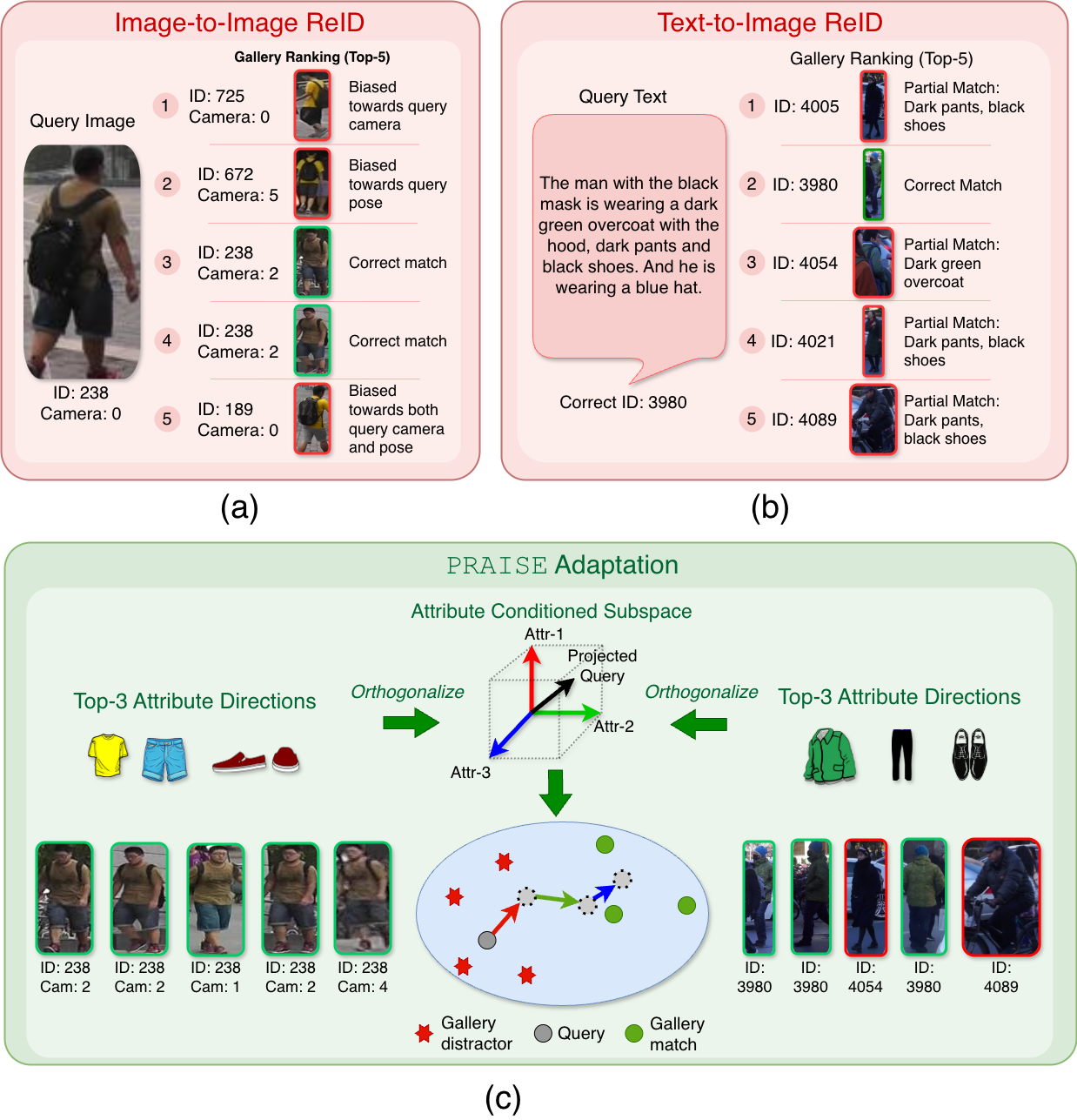}

    \vspace*{-5pt}
    \caption{\textbf{Motivation and overview of} \method{}. (a) In I2I-ReID, raw visual similarity can retrieve distractors caused by camera, pose, or background bias. (b) In T2I-ReID, incomplete text descriptions can produce partial matches that share only some mentioned attributes.  (c) \method{} addresses both cases by selecting query-relevant attribute directions, orthogonalizing them into an Attribute-Conditioned Subspace (ACS), and using this controlled subspace to reshape query-gallery affinities during test-time.}
    \vspace{-18.0pt}
    \label{fig:intro_motivation}
\end{figure}

\vspace{-10.0pt}
\section{Introduction}
\label{sec:intro}
\vspace{-6.0pt}

Person re-identification (ReID) aims to retrieve images of the same identity across cameras and time. It is commonly studied in two retrieval settings: \emph{image-to-image} (I2I) ReID, where an image query is matched against an image gallery, and \emph{text-to-image} (T2I) ReID, where a natural-language description is used to retrieve the corresponding person image. Despite steady progress in both settings using image and text supervision \cite{li2023clip, zuo2024plip}, test-time retrieval remains sensitive to nuisance factors. In I2I-ReID, viewpoint changes, pose variation, blur, illumination, occlusion, and background clutter can make visually plausible distractors outrank the true match \cite{song2025exploring}. In T2I-ReID, retrieval can fail for a complementary reason: natural-language descriptions are often incomplete, ambiguous, or noisy, and may omit discriminative visual evidence \cite{qin2024noisy}. Thus, although the query modality differs, both settings suffer from insufficiently reliable query-gallery evidence at test time.

A natural way to strengthen this evidence is to use semantic attributes as an auxiliary matching signal. If a query and a gallery image depict the same identity, their match should be supported not only by raw embedding similarity, but also by agreement over stable person-centric cues such as clothing color, clothing type, carried objects, hair, or accessories. Conversely, if two samples correspond to different identities, at least some of these cues are likely to disagree. We refer to such cues as \emph{attributes}: attributes that are visually observable, likely to remain meaningful across camera views, and expressible in both image and text queries. These attributes are not intended to uniquely identify a person, but they can provide useful corrective evidence for rejecting implausible matches and reinforcing plausible ones.

Attribute-based cues have been widely explored in person retrieval, including attribute-based ReID and text-based person search \cite{wang2026person, yang2023towards, huang2024attribute, eom2025cerberus}. Prior methods typically use attributes as predicted labels, auxiliary supervision, or learned cross-modal embeddings, relying on training or fine-tuning to calibrate visual evidence with attribute semantics. 
In contrast, our work considers a more realistic setting in which no target-domain identity labels or attribute annotations are available, and no additional attribute classifier can be trained. Under these constraints, we construct a transferable \emph{bank of attributes} that capture visually observable, person-centric cues, such as clothing color and type, hair, and accessories, that remain stable across camera views and can be consistently expressed in both image and text queries.
VLMs \cite{li2023clip,jose2025dinov2,zuo2024plip} can encode these cues as natural-language prompts, allowing textual semantics to be reused in I2I-ReID, where the text branch is typically discarded at inference \cite{li2023clip}, while naturally supporting T2I-ReID. A simple approach is to compare query and gallery responses to this prompt bank and fuse the resulting attribute similarity with identity affinity. However, we show that this direct fusion is unreliable because attribute prompts are coarse, shared across identities, correlated in VLM text space, and biased by prompt templates. Consequently, raw attribute scores are often poorly calibrated and can amplify generic distractors instead of improving retrieval.

This motivates a different test-time use of attributes. Instead of acting as direct matching scores, prompts should guide the adaptation geometry itself. To this end, we propose \method{}: \textbf{PR}ompt-guided \textbf{A}ttribute-based \textbf{I}nference-time \textbf{S}imilarity \textbf{E}nhancement. For each query, \method{} selects the most compatible attribute prompts and orthogonalizes their directions to form an \emph{Attribute-Conditioned Subspace} (ACS). Query-gallery comparisons are then reshaped within this low-rank, query-conditioned subspace, allowing attributes to provide structured auxiliary evidence without relying on brittle prompt-profile matching. The resulting ACS affinity is combined with the standard identity affinity and used to drive a lightweight test-time adaptation objective that updates only normalization parameters. We assume a transductive retrieval setting where the gallery is available at test time, following prior works \cite{adachi2024test, wang2020tent}. The same mechanism applies to both I2I-ReID and T2I-ReID, differing only in whether the query representation is produced by the visual or text encoder.

\noindent \textbf{Our contributions are summarized as follows:}
\noindent \textbf{(1)} \method{}, a test-time adaptation method for ReID that improves query-gallery ranking by combining identity affinity with prompt-guided attribute evidence while updating only lightweight normalization parameters.
\noindent \textbf{(2)} The Attribute-Conditioned Subspace (ACS), a test-time-compatible mechanism that uses attribute prompts as query-conditioned geometric directions rather than direct matching scores, mitigating prompt-similarity bias during retrieval.
\noindent \textbf{(3)} A label-free, transferable attribute prompt-bank construction protocol for ReID. The bank captures stable person-centric cues, is pre-encoded once, requires no target-domain identity or attribute labels, and supports both I2I and T2I retrieval.

\vspace*{-8pt}
\section{Related Works}
\label{sec:survey}
\vspace*{-6pt}

\mypar{Image-to-Image (I2I) Re-identification.} I2I-ReID methods were developed as the principal form of Re-identification in different scenarios such as ordinary/traditional ReID~\cite{haque2016recurrent, karianakis2018reinforced, yang2019attention, si2018dual}, clothes-changing ReID~\cite{li2024rethinking, li2022cocas+, gu2022clothes, huang2021clothing}, Visual Infrared-ReID~\cite{Wang_2019_ICCV_AlignGAN, kniaz2018thermalgan, all-survey, Bi-Di_Center-Constrained, hetero-center, park2021learning} for low lighting conditions, etc. Recently, models such as CLIP-ReID~\cite{li2023clip} have emerged, which use the superior representation power of VLMs in order to perform Re-identification. General-purpose models such as PLIP~\cite{zuo2024plip} have also been proposed, which are essentially VLMs intended for use in ReID settings for both I2I and T2I (Text-to-image) ReID. However, existing VLM-based ReID methods largely underutilize multimodal alignment by relying on cosine similarity between visual embeddings. We argue that this image-only approach overlooks the text branch as a stable semantic prior for mitigating domain shifts, such as camera bias, pose variation~\cite{song2025exploring}, and visual noise.

\mypar{Test-time Adaptation of VLMs.} Several methods have been proposed to adapt VLMs to unseen data streams. CLIPArTT~\cite{hakim2024clipartt} introduces a cross-modal pseudo-labeling loss that refines image-wise prompts, while WATT~\cite{osowiechi2024watt} utilizes prompt ensembling to stabilize retrieval. However, these loss functions often remain heuristically decoupled from the original contrastive geometry of the backbone. Alternative paradigms, such as TPT~\cite{shu2022test}, employ prompt tuning to optimize context tokens at test time, but reliance on multiple stochastic augmentations introduces significant computational overhead. Gradient-free alternatives like TDA~\cite{karmanov2024efficient} utilize cache-based pseudo-labeling to simulate few-shot episodes, yet struggle under severe out-of-distribution shifts where the cached support becomes unreliable. In contrast to classification-centric TTA methods, we propose \method{}, which modulates adaptation geometry through an attribute-conditioned subspace. By updating only a small set of normalization parameters, \method{} enables robust open-set adaptation without unstable global updates or costly augmentation-based tuning.

\mypar{Text-to-Image (T2I) Re-identification.} Existing T2I-ReID methods~\cite{bai2023rasa, jiang2023cross, qin2024noisy, tan2024harnessing, zhu2021dssl, wang2020vitaa, wu2021lapscore} predominantly follow a tripartite taxonomy based on architectural design: (i) unimodal alignment via modality-specific encoders (e.g., ResNet~\cite{he2016deep}, BERT~\cite{wang2024spectrum}), (ii) zero-shot retrieval using general-purpose vision-language models like CLIP~\cite{radford2021learning}, and (iii) domain-specific multimodal pre-training tailored for ReID. While recent advancements in cross-modal embedding models have bridged the high-level semantic gap, these offline frameworks remain constrained by a static internal knowledge base, making them brittle against dynamic or underspecified text queries during inference. Recent literature has explored interactive paradigms, such as THI~\cite{qin2025human}, which utilizes multi-round visual question answering with MLLMs to refine query intent. However, these methods often incur high computational cost and suffer from label leakage or error accumulation. In contrast, \method{} uses an attribute-conditioned subspace (ACS), aligning visual features with textual prompts via lightweight normalization updates. This enables efficient open-set adaptation without multi-round refinement.

\vspace{-4.0pt}
\section{Background: Attribute-Guided Retrieval}
\label{sec:background}
\vspace{-6.0pt}

\mypar{Notation.} Denoting as $\DD_g\,{=}\,\{(x_j,p_j,c_j)\}_{j=1}^{N_g}$ the gallery where $x_j$ is a gallery image, $p_j$ is a person label and $c_j$ is a camera label\footnote{The labels are only denoted for completeness, and are only used for evaluation}, we precompute gallery embeddings: $\ZZ_{g}\,{=}\,\venc(\XX_g)\in\Real^{N_g\times d}$, where each row $\zz_{g,j}$ is $\ell_2$-normalized. In practice, $\ZZ_{g}$ is cached offline, once per test set, to avoid excessive computational loads. Next, let $\PP\,{=}\,\{s_\ell\}_{\ell=1}^{L}$ be a bank of attribute prompts spanning multiple attribute categories. The specific design of the prompt bank will be discussed later. Encoded prompts are denoted as $\ZZ_t = \tenc(\PP)\in\Real^{L\times d}$ with a row-wise $\ell_2$ normalization.

The central goal of ReID is to rank a gallery for a given query such that images of the same identity appear near the top of the list. In most modern VLM-based pipelines, this is done by comparing normalized embeddings from a visual encoder $\venc$ and/or text encoder $\tenc$. For a query ($x_i$, image or text), the query representation is obtained as $\zz_i=\venc(x_i)$ or $\zz_i=\tenc(x_i)$ depending on query modality. Gallery images are encoded as $\ZZ_g=\venc(\XX_g)$, where each row $\zz_{g,j}$ corresponds to one gallery image. The standard retrieval affinity is then computed by cosine similarity:
\vspace{-3.0pt}
\begin{equation}
    \idmx[i,j] = \tilde{\zz}_i^\top \tilde{\zz}_{g,j},
    \label{eq:raw_score}
\end{equation}
\vspace{-3.0pt}
where $\tilde{\zz}=\zz/(\|\zz\|_2+\epsilon)$. This identity score is simple and effective, but it is also vulnerable to nuisance factors such as viewpoint, illumination, camera style, background clutter, occlusion, and incomplete textual descriptions. These factors can cause visually plausible distractors to receive high similarity even when they do not share the same identity.

\vspace{-4.0pt}
\subsection{Attribute Scores as Corrective Retrieval Signal}
\label{sec:bg_score_fusion}
\vspace{-4.0pt}

A natural way to improve retrieval robustness is to introduce semantic attributes as an additional source of evidence. Attributes such as clothing color, sleeve length, lower-body clothing type, gender, hair, carried objects, and accessories are not sufficient to uniquely identify a person, but they often provide useful constraints on whether a query-gallery pair is plausible. Following this intuition, an established design pattern is \emph{score fusion} \cite{aggarwal2020text, lee2025ra}. The retrieval system computes two affinities for each query-gallery pair: an identity or appearance affinity $\idmx$, and an attribute-based affinity $\attmx$. The final matching score is then formed as a weighted combination:
\vspace{-3.0pt}
\begin{equation}
    \bAA[i,j] = \idmx[i,j] + \lambda_{\mathrm{att}} \attmx[i,j],
    \label{eq:bg_score_fusion}
\end{equation}
\vspace{-3.0pt}
where $\lambda_{\mathrm{att}}\ge 0$ controls the contribution of the attribute branch. Conceptually, $\idmx$ captures the raw embedding-level similarity between the query and gallery image, while $\attmx$ acts as a corrective semantic score. The identity score can come from an existing ReID backbone, while the attribute score can be computed using an auxiliary attribute classifier, a text-attribute model, or a VLM prompt bank. A high attribute score should promote candidates that share query-consistent semantic evidence, while a low attribute score should suppress candidates whose appearance similarity is likely caused by nuisance correlations.

Let $\PP\,{=}\,\{s_\ell\}_{\ell=1}^{L}$ denote a bank of natural-language attribute prompts, where each prompt describes one semantic attribute, for example \texttt{``a photo of a person wearing a red shirt''} or \texttt{``a photo of a person carrying a backpack''}. These prompts are encoded by the text encoder:
\begin{equation}
    \ZZ_t = \tenc(\PP) \in \Real^{L\times d},
    \label{eq:bg_prompt_bank}
\end{equation}
with row-wise $\ell_2$ normalization to get $\bar{\ZZ}_t$. For a query embedding $\zz_i$, its similarity to the attribute bank is $\mathbf{u}_i = \bar{\ZZ}_t \zz_i^\top \in \Real^{L}$. Similarly, for a gallery embedding $\zz_{g,j}$, its attribute similarity profile is $\mathbf{v}_j = \bar{\ZZ}_t \zz_{g,j}^\top \in \Real^{L}$. The vectors $\mathbf{u}_i$ and $\mathbf{v}_j$ can be interpreted as soft attribute profiles. Each dimension measures how compatible the query or gallery sample is with one prompt in the attribute bank. Therefore, if the query and gallery image correspond to the same identity, one may expect their attribute profiles to be similar. This leads to a direct attribute similarity:
\vspace{-3.0pt}
\begin{equation}
    \attmx^{\text{direct}}[i,j]
    =
    \tilde{\mathbf{u}}_i^\top \tilde{\mathbf{v}}_j,
    \label{eq:bg_direct_attr_score}
\end{equation}
\vspace{-3.0pt}
where $\tilde{\mathbf{u}}_i\,{=}\,\mathbf{u}_i/(\|\mathbf{u}_i\|_2+\epsilon)$ and $\tilde{\mathbf{v}}_j\,{=}\,\mathbf{v}_j/(\|\mathbf{v}_j\|_2+\epsilon)$. In this view, $\attmx^{\text{direct}}[i,j]$ measures whether the query and gallery induce similar responses over the same set of attribute prompts. It is a query-gallery similarity computed not in the original image/text embedding space, but in the prompt-response space.

The appeal of this construction is that it converts attributes into a retrieval-compatible score without requiring explicit attribute labels. It also applies to both I2I and T2I ReID. In I2I ReID, both $\mathbf{u}_i$ and $\mathbf{v}_j$ are obtained by comparing image embeddings to text prompts. In T2I ReID, $\mathbf{u}_i$ is obtained from a text-query embedding and $\mathbf{v}_j$ from a gallery-image embedding, but both are still expressed as responses to the same prompt bank. Thus, direct attribute similarity provides a unified mechanism for using semantic attributes across both retrieval settings.

\vspace{-4.0pt}
\subsection{Calibration of Prompt-Response Similarity}
\label{sec:bg_pitfall_direct_attr}
\vspace{-4.0pt}

Although direct attribute similarity offers a simple auxiliary retrieval score, it assumes that prompt activations form a reliable matching space. This is unsuitable for ReID because: (1) \textbf{Identity Closeness:} coarse attributes (e.g., ``dark hair'') are shared by many identities and are not identity-specific; (2) \textbf{Template Bias:} responses may reflect shared prompt templates rather
than visual evidence; e.g., \emph{``a photo of a person wearing a red shirt''} and \emph{``a photo of a person wearing a blue shirt''} differ only in the color token but share almost the entire sentence structure. \cite{allingham2023simple,parisot2023learning}  (3) \textbf{Attribute Collision:} semantically close captions can produce highly correlated activations; e.g., \emph{red}/\emph{pink}/\emph{magenta}, \emph{pants}/\emph{trousers}/\emph{long pants}, or \emph{bag}/\emph{backpack}/\emph{shoulder bag} \cite{jung2025learning,tian2024argue}. As a result, $\attmx^{\text{direct}}$ may have a narrow or poorly structured distribution, making true matches and distractors hard to separate and misaligning it with the raw identity affinity $\idmx$.

\begin{figure}[t]
    \centering
    \begin{subfigure}[t]{0.48\columnwidth}
        \centering
        \includegraphics[width=\linewidth]{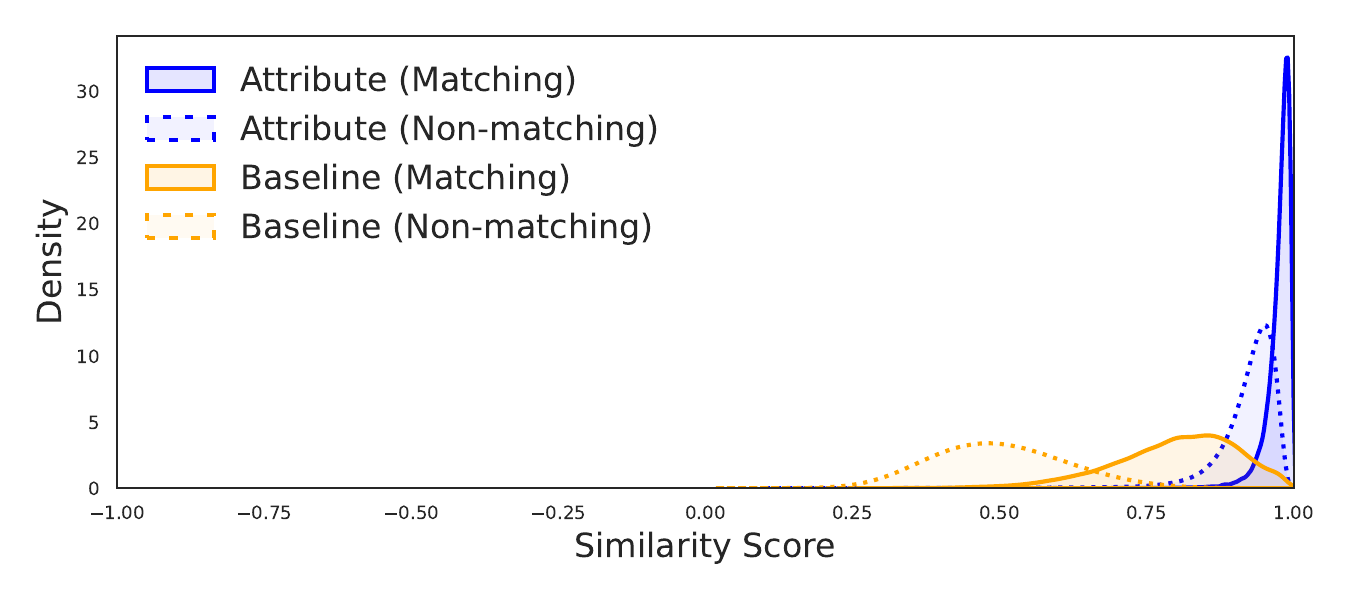}
        \caption{Without ACS}
        \label{fig:ablation_alpha}
    \end{subfigure}\hfill
    \begin{subfigure}[t]{0.48\columnwidth}
        \centering
        \includegraphics[width=\linewidth]{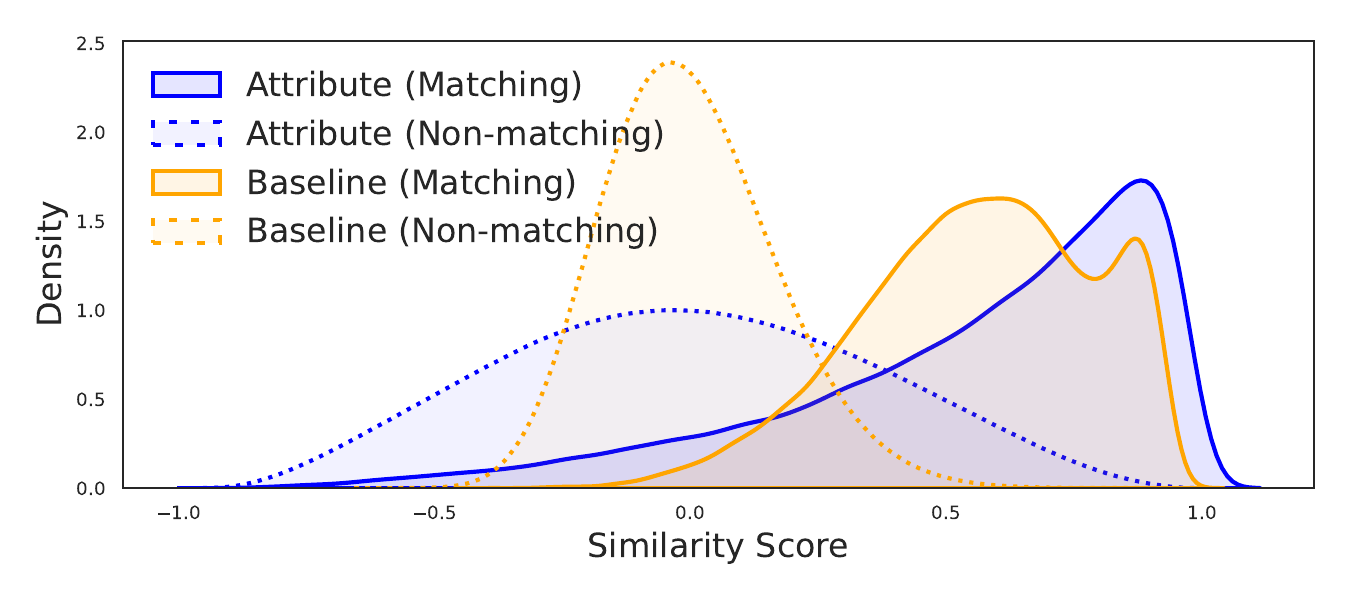}
        \caption{With ACS}
        \label{fig:ablation_topk}
    \end{subfigure}

    \vspace*{-5pt}
    \caption{Similarity distributions for matching and non-matching query-gallery pairs on Market-1501. Direct prompt-profile similarity (a) is saturated for both groups, whereas ACS (b) yields clearer separation between matches and distractors.}
    \label{fig:hist_row}
    \vspace{-16.0pt}
\end{figure}

This behavior is illustrated in \autoref{fig:hist_row}. The direct attribute-derived affinity exhibits a markedly different distribution from the baseline identity affinity, indicating weak agreement between the two matching views. Under such mismatch, additive fusion becomes unstable: the attribute term is intended to correct $\idmx$, but can instead amplify distractors that merely share generic attributes with the query.

This suggests that attribute prompts should not be used as unconstrained scoring functions. Instead, they should define a more controlled comparison space in which identity-level affinity remains the primary signal, while prompt-template bias is reduced and correlated attribute captions are prevented from dominating the score. Such a formulation would preserve the corrective value of attributes without allowing coarse or colliding prompt responses to destabilize retrieval.

\vspace{-4.0pt}
\section{Attribute conditioned subspace construction and \method{}}
\label{sec:methods}
\vspace{-6.0pt}

\begin{figure*}[!ht]
    \centering
    \includegraphics[width=0.8\linewidth]{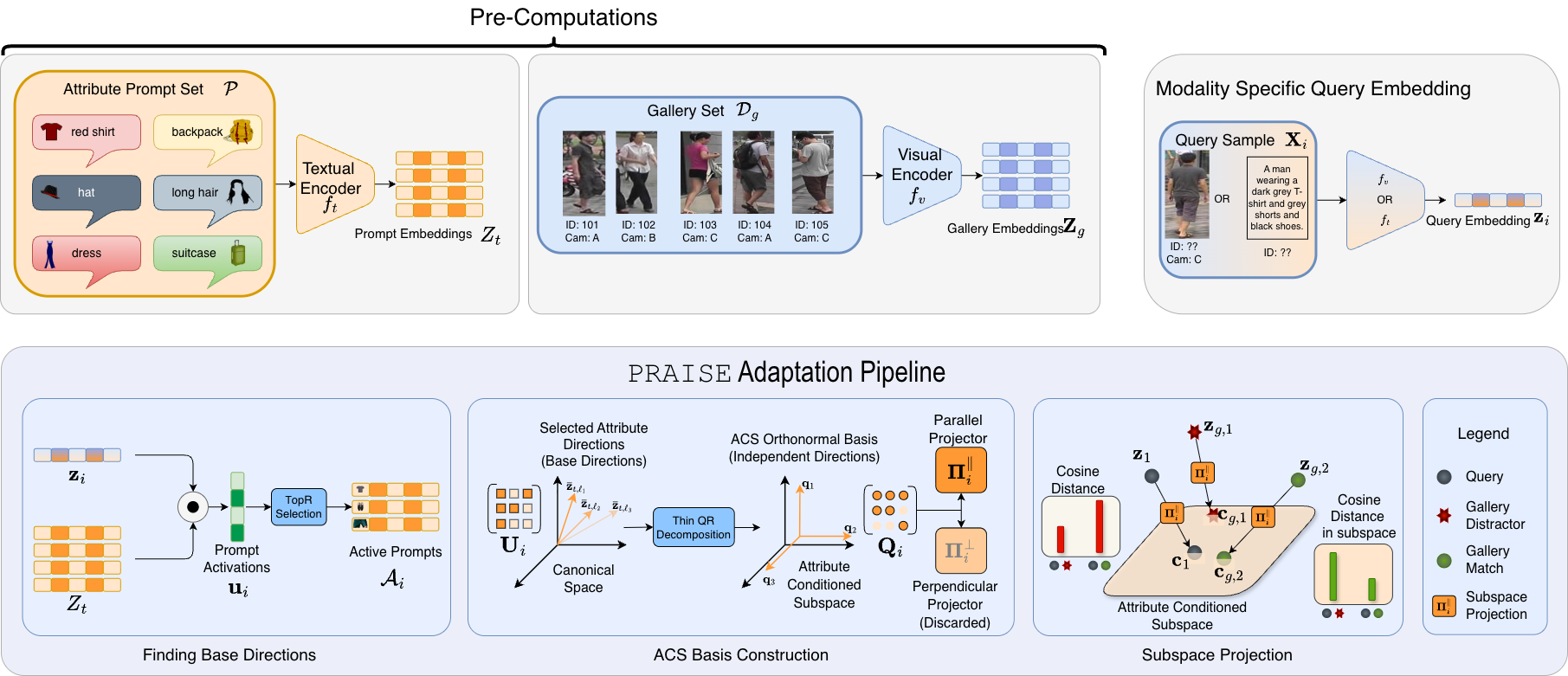}

    \vspace*{-5pt}
    \caption{\textbf{Method overview.} \method{} uses attribute prompts to construct a query-specific semantic subspace that stabilizes query-side adaptation under domain shift. At test time, it performs query-side episodic TTA with a fixed source-space gallery index. Gallery and prompt-bank embeddings are computed once and kept fixed; gradients update only the query-side normalization affine parameters.}
    \label{fig:method-figure}
    \vspace{-12.0pt}
\end{figure*}

Given a frozen VLM and a target-domain test set (queries and gallery), we propose \method{}, a TTA method that leverages noisy natural-language attribute prompts to recalibrate the embedding space under domain shift to improve query-gallery ranking for both I2I-ReID and T2I-ReID. \method{} is a TTA method that performs the following key steps: (1) precompute a bank of attribute prompt embeddings, (2) construct a \emph{query-conditioned attribute subspace} using the prompt bank, and (3) adapt a small subset of backbone parameters at test time, to improve retrieval performance. Please refer to \autoref{fig:method-figure} for an overview of the methodology.

\vspace{-4.0pt}
\subsection{Prompt Bank Construction}
\label{sec:prompt_bank}
\vspace{-4.0pt}


The quality of the attribute bank is important because \method{} relies on prompt-induced semantic directions rather than supervised attribute labels. We therefore construct $\PP$ through a data-driven but label-free procedure that extracts recurring pedestrian attributes while keeping the vocabulary compact and reproducible.

\begin{itemize}
    \item \textbf{Captioning.} We use Qwen2.5-VL-7B-Instruct~\cite{yang2025qwen3}
    to generate short person-centric descriptions for pedestrian images. The dataset can be any standard ReID dataset. These descriptions are used only to mine recurring attributes, not as identity annotations.

    \item \textbf{Attribute grouping.} We map the generated descriptions into coarse pedestrian categories, including hair, gender, upper-/lower-body clothing color, clothing type, and accessories.

    \item \textbf{Vocabulary cleaning.} Within each category, we keep unique attribute phrases and remove duplicate or near-identical variants, yielding a compact set of attribute suffixes. Each suffix is inserted into the shared template $s_\ell=\texttt{``A photo of [ATT]''}$ to form
    $\PP=\{s_\ell\}_{\ell=1}^{L}$. The full suffix list is provided in the Supplementary Material.

    \item \textbf{Prompt encoding and centering.} We encode the bank as $\ZZ_t=\tenc(\PP)\in\Real^{L\times d}$ and apply row-wise $\ell_2$ normalization. To reduce shared-template bias, we also encode a neutral prompt $s_0$, e.g., \emph{``a photo of a person''}, and use centered directions $\bar{\zz}_{t,\ell}=\zz_{t,\ell}-\zz_{t,0}$ for $\ell\in\{1,\dots,L\}$.
\end{itemize}

Unless otherwise stated, we mine the attribute bank once from the Market-1501 training split and freeze it for all experiments. The same centered bank $\bar{\ZZ}_t$ is reused across all evaluation datasets and is not adapted using target identity labels, target attribute labels, or dataset-specific supervision. The resulting bank should therefore be understood as a fixed, label-free, transferable pedestrian attribute vocabulary rather than a dataset-specific classifier. In the Supplementary Material, we report prompt-bank source ablations showing that \method{} remains stable when the bank is mined from different ReID datasets.

\vspace{-4.0pt}
\subsection{Attribute-Conditioned Subspace (ACS) Construction}
\label{sec:method:acs}
\vspace{-4.0pt}
We now define the attribute-conditioned subspace (ACS) that will be used by \method{}. For a given query, ACS is a low-rank, query-conditioned subspace spanned by the text directions of the most compatible attribute prompts. Rather than treating prompt similarities as final matching scores, ACS uses prompts to identify a small set of semantic directions along which query-gallery comparisons can be made more reliably. Thus, the prompt bank acts as the source of candidate attribute directions from which the query-specific subspace is constructed. 

\mypar{Finding the base attribute directions.}
For each query embedding $\zz_i$ (row $i$ of $\ZZ$), we compute prompt activations using the prompt bank: $\mathbf{u}_i \,=\, \bar{\ZZ}_t\,\zz^\top_i\in\Real^{L}$. Please note that the query embedding ($\ZZ$) is the encoder representation of a query image batch ($\XX_q$) in case of I2IReID (i.e., $\venc(\XX_q)$) or a query text description ($\mathcal{P}_q$) in case of T2IReID (i.e., $\tenc(\mathcal{P}_q)$). We then select a small index set of the most activated prompts:
\vspace{-3.0pt}
\[
\bA_i \,=\, \operatorname{TopR}(\mathbf{u}_i),\qquad |\bA_i|=r,\qquad r\ll L.
\]
\vspace{-3.0pt}
This selection is treated as an index operation, i.e., stop-gradient through $\operatorname{TopR}$. This allows us to find a set of attribute direction vectors: 
\vspace{-3.0pt}
\[
\mathbf{U}_i \,=\, 
\left[\bar{\zz}_{t,\ell_1} \cdots \,\bar{\zz}_{t,\ell_r}\right]^\top
\in\Real^{r\times d},
\qquad \ell_k\in\bA_i.
\]
\vspace{-3.0pt}

\vspace{-3.0pt}
\mypar{Constructing the query-conditioned basis.}
While centering reduces template bias, attribute collision remains unresolved: selected prompt directions can still be redundant or unevenly scaled, causing correlated attributes to over-count the same cue. We therefore orthonormalize these directions into a controlled basis, so that projection-based comparisons are less sensitive to redundancy and each axis contributes independently. This is done via thin QR decomposition as follows:
\vspace{-3.0pt}
\[
\mathbf{U}_i^\top \,=\, \QQ_i\,\mathbf{R}_i,\qquad
\QQ_i\in\Real^{d\times m},\quad
\QQ_i^\top\QQ_i=I_m,\quad m\le r.
\]
\vspace{-3.0pt}
We refer to $\QQ_i$ as the \emph{Attribute-Conditioned Subspace (ACS) basis} for query $i$.

\mypar{Projection operators.}
$\QQ_i$ can now be decomposed into two sets of projection operators, one component parallel to the direction vectors and the other one perpendicular. We define these subspace projectors as: $\PPi_i^{\parallel} \,=\, \QQ_i\QQ_i^\top$ and $\PPi_i^\perp \,=\, I-\PPi_i^{\parallel}$. We only need the subspace parallel projections and therefore, the projected features can be written as: $\zz_i^{\parallel}=\PPi_i^{\parallel}\zz_i$ for the query and, $\zz_{g,j}^{\parallel}=\PPi_i^{\parallel}\zz_{g,j}$ for the gallery. Note that we will use the $\ell_2$-norm forms of the projected features frequently from this point onwards, and we denote them as: $\tilde{\zz} \,=\, \frac{\zz}{\|\zz\|_2+\epsilon}$. This makes computation of cosine similarities much easier by using only an inner product.

\emph{Implementation Note:}
We never explicitly form $\PPi_i^{\parallel}\in\Real^{d\times d}$. Given $\QQ_i\in\Real^{d\times k_i}$ with orthonormal columns, projection is applied as: $\PPi_i^{\parallel} \zz = \QQ_i(\QQ_i^\top \zz)$ and $\PPi_i^\perp \zz = \zz - \QQ_i(\QQ_i^\top \zz)$, which costs $\mathcal{O}(dk_i)$ rather than $\mathcal{O}(d^2)$.\footnote{\emph{Equivalence:}
Since $\QQ_i^\top\QQ_i=I_{k_i}$, for any vectors $\zz_a,\zz_b\in\Real^d$: $(\PPi_i^{\parallel} \zz_a)^\top(\PPi_i^{\parallel} \zz_b) = (\QQ_i^\top \zz_a)^\top(\QQ_i^\top \zz_b)$ such that $\|\PPi_i^{\parallel} \zz_a\|_2 = \|\QQ_i^\top \zz_a\|_2$.} The full process is summarized in Algorithm \ref{alg:acs_efficient}. 

\mypar{ACS score computation.}
We now define the subspace (attribute) affinity as a measure of alignment towards the specific set of direction vectors:
\vspace{-3.0pt}
\begin{equation} \label{eq:att_afinity}
    \bAA^{\parallel}[i,j] \,=\, (\tilde{\zz}_i^{\parallel})^\top \tilde{\zz}_{g,j}^{\parallel}.
\end{equation}
\vspace{-3.0pt}

In practice, this can be computed simply using $\mathbf{c}_i=\zz_i^\top\QQ_i$ and $\mathbf{c}_{g,j}=\zz_{g,j}^\top\QQ_i$, $\ell_2$-normalizing them, and then taking a dot product. We then use a single scalar $\alpha_{\mathrm{acs}}\ge 0$ to blend the identity affinity with the attribute-subspace affinity. The ACS score is therefore derived as a linear combination of both affinities. The final ACS affinity used for computing inference time loss is:  

\vspace{-3.0pt}
\begin{equation} \label{eq:final_acs_affinity}
    \bAA[i,j] \,=\, \idmx[i,j] \,+\, \alpha_{\mathrm{acs}}\,\bAA^{\parallel}[i,j].
\end{equation}
\vspace{-3.0pt}

\begin{algorithm}[t]
{
\footnotesize
\caption{Efficient ACS affinity computation}
\label{alg:acs_efficient}
\begin{algorithmic}[1]
\Require Normalized query embedding $\tilde{\zz}_i$, normalized gallery embeddings $\tilde{\ZZ}_g$, centered prompt bank $\bar{\ZZ}_t$, selection size $r$
\Ensure ACS affinity scores $\bAA^{\parallel}[i,:]$

\State Query-attribute similarities: $\mathbf{u}_i\,{\gets}\,\bar{\ZZ}_t \zz_i^\top$
\State Top-r indexing: $\bA_i\,{\gets}\,\operatorname{TopR}(\mathbf{u}_i)$, with $|\bA_i|=r$
\State Forming direction vectors: $\mathbf{U}_i\,{\gets}\,[\bar{\zz}_{t,\ell_1}\ \cdots\ \bar{\zz}_{t,\ell_r}]^\top$, $\ell_m\in\bA_i$
\State Orthogonalization: $\mathbf{U}_i^\top = \QQ_i\mathbf{R}_i$, with $\QQ_i^\top\QQ_i=\mathbf{I}$

\State Compute ACS projections: $\mathbf{c}_i\,{\gets}\,\zz_i^\top\QQ_i$ and $\mathbf{C}_g\,{\gets}\,\ZZ_g\QQ_i$

\State Normalize projections: $\tilde{\mathbf{c}}_i\,{\gets}\,\frac{\mathbf{c}_i}{\|\mathbf{c}_i\|_2+\epsilon}$ and $\tilde{\mathbf{C}}_g\,{\gets}\,\operatorname{row\text{-}norm}(\mathbf{C}_g)$

\State Compute ACS affinity: $\bAA^{\parallel}[i,:]\,{\gets}\,\tilde{\mathbf{c}}_i \tilde{\mathbf{C}}_g^\top$

\State \Return $\bAA^{\parallel}[i,:]$
\end{algorithmic}
}
\end{algorithm}

\vspace{-8.0pt}
\subsection{Test-Time Adaptation Objective}
\label{sec:method:loss}
\vspace{-4.0pt}

\mypar{Retrieval distribution.}
Given logits $\bAA\in\Real^{B\times N_g}$, we define a retrieval distribution per query: $\mathbf{p}_i \,=\, \operatorname{softmax}(\bAA[i,:]/\tau)$, where $\tau>0$ is a temperature value that controls the shape of the softmax distribution. Gallery and prompt-bank embeddings are kept fixed as source-space anchors and only the query-side encoder path is updated.

\mypar{Top-$k$ candidate-set likelihood.}
Let $\mathcal{K}_i=\operatorname{TopK}(\mathbf{p}_i)$ denote indices of top-$k$ probabilities.
We minimize a top-$k$ negative log-likelihood:

\vspace{-6.0pt}
\begin{equation}
    \mathcal{L}_{\text{TTA}}
    \,=\,
    -\frac{1}{B}\sum_{i=1}^{B}\sum_{j\in\mathcal{K}_i}\log \mathbf{p}_i(j).
\end{equation}

This encourages the retrieval distribution to become concentrated on a small candidate set, which empirically stabilizes adaptation under large galleries. Please refer to the Supplementary Materials for a theoretically grounded motivation for the loss function.

\mypar{Adaptation settings.}
We adapt only normalization parameters (affine scale and bias) within $\venc$: $\,\theta\,{=}\,\{\text{norm affine parameters in }\venc\}$, keeping all other weights frozen. This preserves pretrained semantics and reduces overfitting at test time. We adopt an episodic TTA, i.e., for each test batch, we reset $\theta\leftarrow\theta_0$ (pretrained), perform $T$ gradient steps on $\mathcal{L}$, and output the adapted embeddings for evaluation.  Note that in case of Text-to-Image ReID, only the adaptation parameters become $\theta_t=\{\text{norm affine parameters in }\tenc\}$. The rest of the pipeline follows as is, with the same logic.
\vspace{-5.0pt}
\section{Results and Discussion}
\label{sec:results}
\vspace{-4.0pt}

\subsection{Experimental Protocol}
We evaluate \method{} in two different settings: Image-to-Image (I2I) ReID and Text-to-Image (T2I) ReID. We use MSMT17 \cite{wei2018person} and Market-1501 \cite{zheng2015scalable} for I2I, and CUHK-PEDES \cite{li2017person} and RSTP-ReID \cite{zhu2021dssl} for T2I. At test time, all methods are given (i) the query (image or text) and (ii) the \emph{unlabeled} target gallery. We pre-encode all gallery features once and reuse them for all methods for efficiency. Furthermore, we ensure that all methods are granted the same access to unlabeled gallery features under the same episodic/reset protocol as our approach. All runtimes are measured on an NVIDIA RTX A6000 GPU (single GPU) with identical implementation settings (batch size, mixed precision), and exclude the one-time cost of gallery feature pre-encoding. Adam optimizer has been used throughout, for all experiments.

\vspace{-4.0pt}
\subsection{Impact on different backbones}
\vspace{-4.0pt}

\begin{table}[t]
\centering
\caption{\textbf{Effect on different backbones.} Backbones evaluated on I2IReID and T2IReID datasets with \method{} applied (score on left side of `$/$' is without adaptation and score on right is after applying \method{}). Each dataset reports mAP and Rank-1 Accuracy (R1).}
\label{tab:backbones}

\vspace*{-5pt}
\setlength{\tabcolsep}{5pt}
\renewcommand{\arraystretch}{1.15}
\resizebox{0.9\columnwidth}{!}{%
\begin{tabular}{l | cc | cc !{\vrule width 1.2pt} cc | cc}
\toprule
\multirow{3}{*}{\textbf{Backbone}} 
& \multicolumn{4}{c!{\vrule width 1.2pt}}{\textbf{I2IReID}} 
& \multicolumn{4}{c}{\textbf{T2IReID}} \\
& \multicolumn{2}{c|}{\textbf{MSMT17}} 
& \multicolumn{2}{c!{\vrule width 1.2pt}}{\textbf{Market-1501}} 
& \multicolumn{2}{c|}{\textbf{CUHK-Pedes}} 
& \multicolumn{2}{c}{\textbf{RSTPReID}} \\
& \textbf{mAP} & \textbf{R1} 
& \textbf{mAP} & \textbf{R1} 
& \textbf{mAP} & \textbf{R1} 
& \textbf{mAP} & \textbf{R1} \\
\midrule
\rowcolor{Lavender!20}
CLIP ViT-B/16 \cite{radford2021learning}  & \phz1.1\,$/$\,\phz1.4 & \phz5.6\,$/$\,\phz6.7 & \phz3.5\,$/$\,\phz4.3 & 11.9\,$/$\,14.0 & \phz7.1\,$/$\,\phz9.1 & \phz8.9\,$/$\,11.7 & \phz8.1\,$/$\,\phz9.2 & 11.2$/$12.9 \\
\rowcolor{Lavender!20}
DINO.txt \cite{jose2025dinov2}  & \phz0.7\,$/$\,\phz1.8 & \phz5.2\,$/$\,\phz9.4 & \phz1.9\,$/$\,\phz3.0 & \phz7.1\,$/$\,10.1 & \phz7.3\,$/$\,\phz8.8 & \phz9.1\,$/$\,12.8 & 10.3\,$/$\,13.1 & 11.9\,$/$\,14.6 \\
\midrule
\rowcolor{ForestGreen!10}
CLIPReID \cite{li2023clip}  & 20.8\,$/$\,24.7 & 45.9\,$/$\,52.8 & 37.8\,$/$\,42.1 & 63.3\,$/$\,69.4 & \phz8.1\,$/$\,\phz9.9 & \phz8.2\,$/$\,11.4 & 12.2\,$/$\,13.8 & 13.5\,$/$\,15.1 \\
\rowcolor{ForestGreen!10}
RDE \cite{qin2024noisy}  & - & - & - & - & 33.5\,$/$\,38.1 & 36.8\,$/$\,42.8 & 41.2\,$/$\,46.3 & 54.9\,$/$\,60.9 \\
\rowcolor{ForestGreen!10}
ICL \cite{qin2025human}  & - & - & - & - & 26.8\,$/$\,30.0 & 41.0\,$/$\,47.4 & 37.5\,$/$\,42.1 & 62.1\,$/$\,66.3 \\
\midrule
\colrow
PLIP \cite{zuo2024plip}          & 10.5\,$/$\,18.2 & 34.2\,$/$\,38.5 & 35.0\,$/$\,46.5 & 62.6\,$/$\,71.1 & 46.4\,$/$\,49.6 & 55.9\,$/$\,58.4 & 23.1\,$/$\,27.6 & 34.5\,$/$\,37.6 \\
\bottomrule
\end{tabular}}
\end{table}

We evaluate \method{} across different VLM backbones in \autoref{tab:backbones}. Only CLIP \cite{radford2021learning}, Dino.txt \cite{jose2025dinov2} and PLIP \cite{zuo2024plip} are taken by their off-the-shelf training weights. For the other methods, we avoid fully supervised training on the datasets we test on, in order to effectively evaluate test-time adaptation strategies. This is why, most of the scores on the table appear to be very poor.\footnote{Please refer to the Supplementary Materials for performance on fully supervised backbones} For CLIP-ReID \cite{li2023clip}, RDE \cite{qin2024noisy} and ICL \cite{qin2025human}, we test on the dataset specified in the column header, upon training on the other dataset in the same task category (I2I or T2I). Off-the-shelf CLIP \cite{radford2021learning} and Dino.txt \cite{jose2025dinov2} perform poorly on ReID and are clearly unsuitable for a meaningful evaluation of \method{}. We find that \method{} consistently improves performance for all the state-of-the-art ReID methods compared. \cite{li2023clip, qin2024noisy, qin2025human}. PLIP \cite{zuo2024plip}, being a general-purpose ReID-oriented VLM, achieves strong performance across both I2I and T2I settings. While \method{} improves each backbone, we use PLIP as the default backbone due to its consistent performance across all evaluated tasks.

\vspace{-8.0pt}
\subsection{Performance on Re-identification}
\vspace{-6.0pt}

\mypar{How does \method{} perform on I2I-ReID?}
A consistent trend is that classifier-oriented TTA methods often \emph{degrade} performance compared to no-adapt (e.g., TENT/SHOT/SAR/ETA) \cite{wang2020tent,liang2020shot,niu2023towards,niu2022efficient}, with only minor gains from DeYO \cite{lee2024entropy}. VLM-based TTA methods \cite{shu2022test,karmanov2024efficient} are more stable, while the strongest baselines are ReID-specific methods such as TEMP \cite{adachi2024test} and Norm \cite{song2025exploring}. We find that direct score fusion has some limited semantic improvement but the poor calibration leads to a poorer Rank-1. \method{} substantially improves mAP over all baselines and also yields consistent Rank-1 gains. Although TDA and Norm are faster because they are backpropagation-free, \method{} provides a strong runtime-accuracy tradeoff. Moreover, since Norm is post-hoc, it can be applied on top of \method{} with only a 0.3ms/sample overhead, further improving performance.

\begin{table}[t]
\centering
\caption{Comparison of TTA methods on \textbf{I2IReID} benchmarks using a PLIP \cite{zuo2024plip} backbone. Each dataset reports mAP and Rank-1 Accuracy (R1). 
}

\vspace*{-5pt}
\label{tab:tta_i2ireid}
\setlength{\tabcolsep}{6pt}
\renewcommand{\arraystretch}{1.15}
\resizebox{0.85\columnwidth}{!}{%

\begin{tabular}{l | c !{\vrule width 1.2pt} cc | cc !{\vrule width 1.2pt} r}
\toprule
\multirow{3}{*}{\textbf{TTA Method}}
& \multirow{3}{*}{\textbf{Type}}
& \multicolumn{4}{c!{\vrule width 1.2pt}}{\textbf{Dataset}}
& \multirow{3}{*}{\makecell{\bf Time (ms) \\ \bf {/}\,sample}} \\
& & \multicolumn{2}{c|}{\textbf{MSMT17}}
& \multicolumn{2}{c!{\vrule width 1.2pt}}{\textbf{Market-1501}}
& \\
& & \textbf{mAP} & \textbf{R1}
& \textbf{mAP} & \textbf{R1}
& \\
\midrule
No-adapt (Source) \venue{NeurIPS'24} & -- & 10.5 & 34.2 & 35.0 & 62.6 & \phz1.8 \\
\midrule
TENT \cite{wang2020tent} \venue{ICLR'21} & \multirow{5}{*}{\makecell{Classifier \\logit}} & \phz9.2 & 29.8 & 29.3 & 52.7 & 16.2 \\
SHOT \cite{liang2020shot} \venue{ICML'20} & & \phz9.9 & 30.2 & 28.9 & 56.8 & 63.8 \\
SAR \cite{niu2023towards} \venue{ICLR'23} & & 10.0 & 31.1 & 33.2 & 63.3 & 91.4 \\
ETA \cite{niu2022efficient} \venue{ICML'22} & & \phz9.8 & 30.8 & 32.2 & 58.9 & 47.1 \\
DeYO \cite{lee2024entropy} \venue{ICLR'24} & & 10.8 & 34.9 & 34.2 & 64.0 & 51.2 \\
\midrule
TPT \cite{shu2022test} \venue{NeurIPS'22} & \multirow{2}{*}{\makecell{VLM \\ based}} & 11.7 & 35.2 & 35.9 & 63.2 & 1189.1 \\
TDA \cite{karmanov2024efficient} \venue{CVPR'24} & & 10.7 & 34.2 & 35.6 & 63.1 & \phz3.4 \\
\midrule
TEMP \cite{adachi2024test} \venue{IJCNN'24} & \multirow{3}{*}{\makecell{ReID / \\ retrieval}} & 11.8 & 36.1 & 37.1 & 65.1 & 20.1 \\
Anchor-ReID \cite{almansoori2023anchor} \venue{SCIA'23} & & \phz7.9 & 24.3 & 31.2 & 45.1 & \phz6.8 \\
Norm \cite{song2025exploring} \venue{ICLR'25} & & 14.8 & \ul{40.2} & 41.0 & 67.3 & \phz2.2 \\
\midrule
\colrow
Direct Score Fusion ($\lambda_{att}{=}0.1$) & & 12.8 & 33.0 & 39.2 & 60.3 & 38.1 \\
\colrow
\method{} (Ours) & & \ul{18.2} & 38.5 & \ul{46.5} & \ul{71.1} & 42.8 \\
\colrow
\method{} + Norm & & \bf 19.5 & \bf 40.4 & \bf 48.1 & \bf 72.5 & 43.1 \\
\bottomrule
\end{tabular}
}

\vspace*{-8pt}
\end{table}

\mypar{How does \method{} perform on T2I-ReID?}
A similar trend holds for T2I-ReID: classifier-logit methods such as TENT and SHOT \cite{wang2020tent,liang2020shot} give only marginal gains, while VLM classification-based methods such as TPT, TDA, WATT, CLIPArTT and CLIPTTA \cite{shu2022test,karmanov2024efficient,osowiechi2024watt,hakim2024clipartt,lafon2025cliptta} are often unstable and can degrade retrieval. TCR \cite{li2024test} is the strongest retrieval-specific baseline, but \method{} achieves the best results on both datasets, improving CUHK-Pedes from 47.6/57.0 to 49.6/58.4 and RSTPReID from 25.8/36.6 to 27.6/37.6 in mAP/R1. Although slower than no-adapt and TCR, \method{} remains much faster than augmentation-heavy methods such as TPT and WATT, while providing stronger retrieval accuracy under incomplete or ambiguous text queries. Similar to I2I-ReID, we find direct score fusion has a positive effect on the overall retrieval but leads to the ranking accuracy going down due to its improper calibration.

\begin{table}[t]
\centering
\caption{Comparison of TTA methods on \textbf{T2IReID} benchmarks on a PLIP \cite{zuo2024plip} backbone. Each dataset reports mAP and Rank-1 Accuracy (R1). 
}
\label{tab:tta_t2ireid}

\vspace*{-5pt}
\setlength{\tabcolsep}{6pt}
\renewcommand{\arraystretch}{1.15}
\resizebox{0.85\columnwidth}{!}{%
\begin{tabular}{l | c !{\vrule width 1.2pt} cc | cc !{\vrule width 1.2pt} c}
\toprule
\multirow{3}{*}{\textbf{TTA Method}} &
 \multirow{3}{*}{\textbf{Type}} &
\multicolumn{4}{c!{\vrule width 1.2pt}}{\textbf{Dataset}}
& \multirow{3}{*}{\makecell{\bf Time (ms) \\ \bf {/}\,sample }} \\
& & \multicolumn{2}{c|}{\textbf{CUHK-Pedes}}
& \multicolumn{2}{c!{\vrule width 1.2pt}}{\textbf{RSTPReID}}
& \\
& & \textbf{mAP} & \textbf{R1}
& \textbf{mAP} & \textbf{R1}
& \\
\midrule
No-adapt (Source) \venue{NeurIPS'24} & -- & 46.4 & 55.9 & 23.1 & 34.5 & 12.9 \\
\midrule
TENT \cite{wang2020tent} \venue{ICLR'21} & \multirow{2}{*}{\makecell{Classifier \\logit}} & 46.4 & 56.2 & 23.3 & 34.2 & 81.8 \\
SHOT \cite{liang2020shot} \venue{ICML'20} & & 47.1 & 56.6 & 24.2 & 34.0 & 179.1 \\
\midrule
TPT \cite{shu2022test} \venue{NeurIPS'22} & \multirow{6}{*}{\makecell{VLM \\based}} & 46.5 & 55.9 & 23.1 & 34.6 & 4193.4 \\
TDA \cite{karmanov2024efficient} \venue{CVPR'24} & &  39.2 & 48.9 & 16.3 & 26.8 & 29.8 \\
WATT\,(S) \cite{osowiechi2024watt} \venue{NeurIPS'24} & & 39.6 & 36.8 & 20.0 & 27.5 & 1339.3 \\
WATT\,(P) \cite{osowiechi2024watt} \venue{NeurIPS'24} & & 40.1 & 47.6 & 20.2 & \phz7.8 & 1398.8 \\
CLIPArTT \cite{hakim2024clipartt}  \venue{WACV'25} & & 37.5 & 44.8 & 20.0 & 27.9 & 605.2 \\
CLIPTTA \cite{lafon2025cliptta}   \venue{NeurIPS'25} & & 41.4 & 49.4 & 22.1 & 31.7 & 592.8 \\
\midrule
TCR \cite{li2024test} \venue{ICLR'25} & Retrieval & 47.6 & 57.0 & 25.8 & 36.6 & 120.8 \\
\midrule
\colrow
Direct Score Fusion ($\lambda_{att}{=}0.1$) & & 48.8 & 52.1 & 24.5 & 32.9 & 132.4 \\
\colrow
\method{} (Ours) & & \bf 49.6 & \bf 58.4 & \bf 27.6 & \bf 37.6 & 160.3 \\
\bottomrule
\end{tabular}}

\vspace*{-8pt}
\end{table}

\noindent \textit{Note.} We find that our prompt bank construction strategy is transferable to other datasets as well, with minimal performance changes when a differently sourced prompt bank is used. Furthermore, we find that the traditional test time adaptation methods are more suited to operating under corruption and image-noise based shifts where entropy based sharpening improves the retrieval geometry. Due to space restrictions in the main manuscript, \emph{a detailed summary with results can be found in the Supplementary Materials.}

\begin{figure*}[!ht]
    \centering
    \begin{subfigure}[t]{0.18\linewidth}
        \centering
        \includegraphics[width=0.9\linewidth]{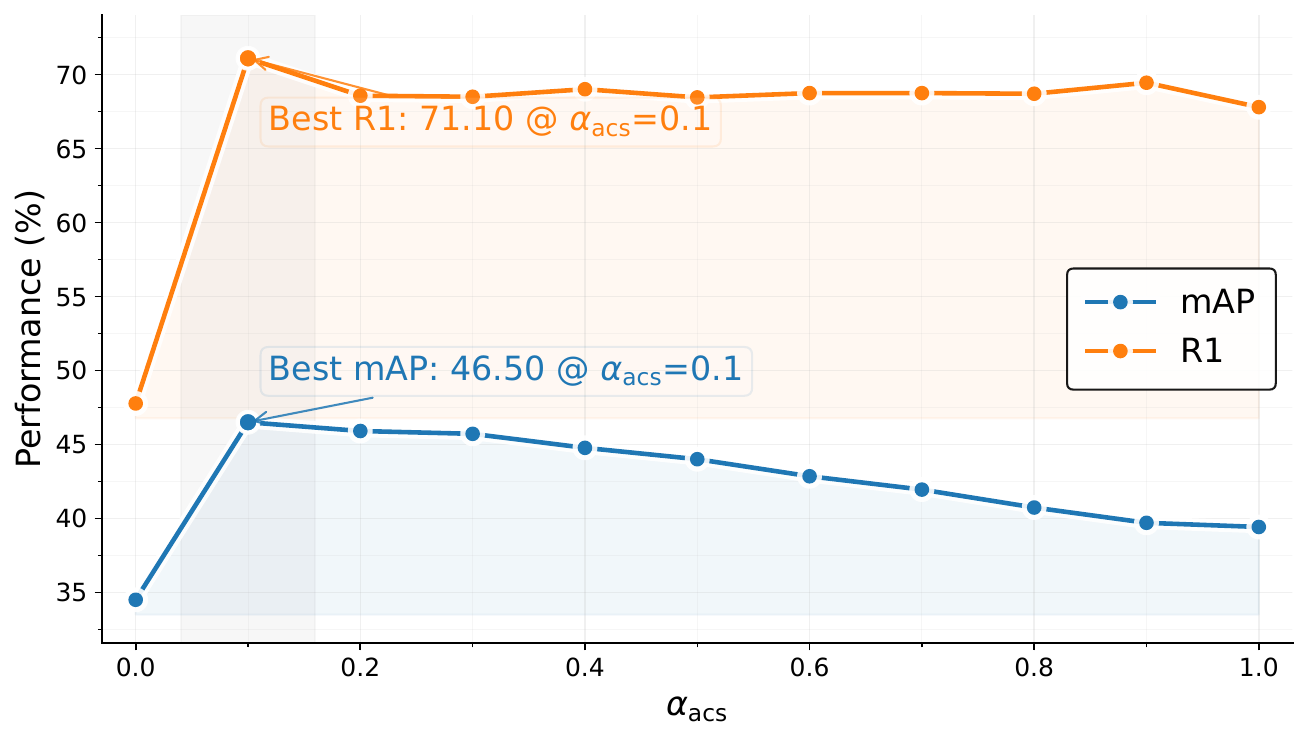}
        \caption{$\alpha_{\mathrm{acs}}$}
        \label{fig:ablation_alpha}
    \end{subfigure}\hfill
    \begin{subfigure}[t]{0.18\linewidth}
        \centering
        \includegraphics[width=0.9\linewidth]{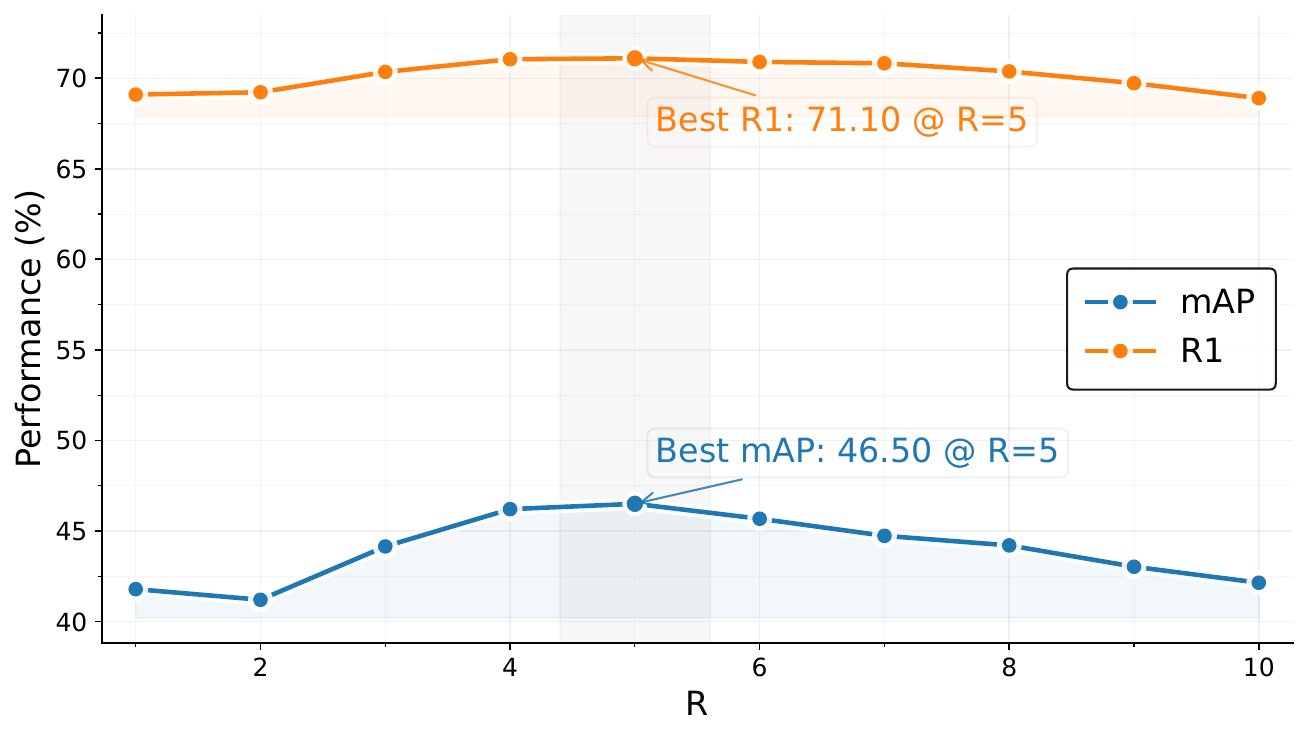}
        \caption{TopR}
        \label{fig:ablation_topr}
    \end{subfigure}\hfill
    \begin{subfigure}[t]{0.18\linewidth}
        \centering
        \includegraphics[width=0.9\linewidth]{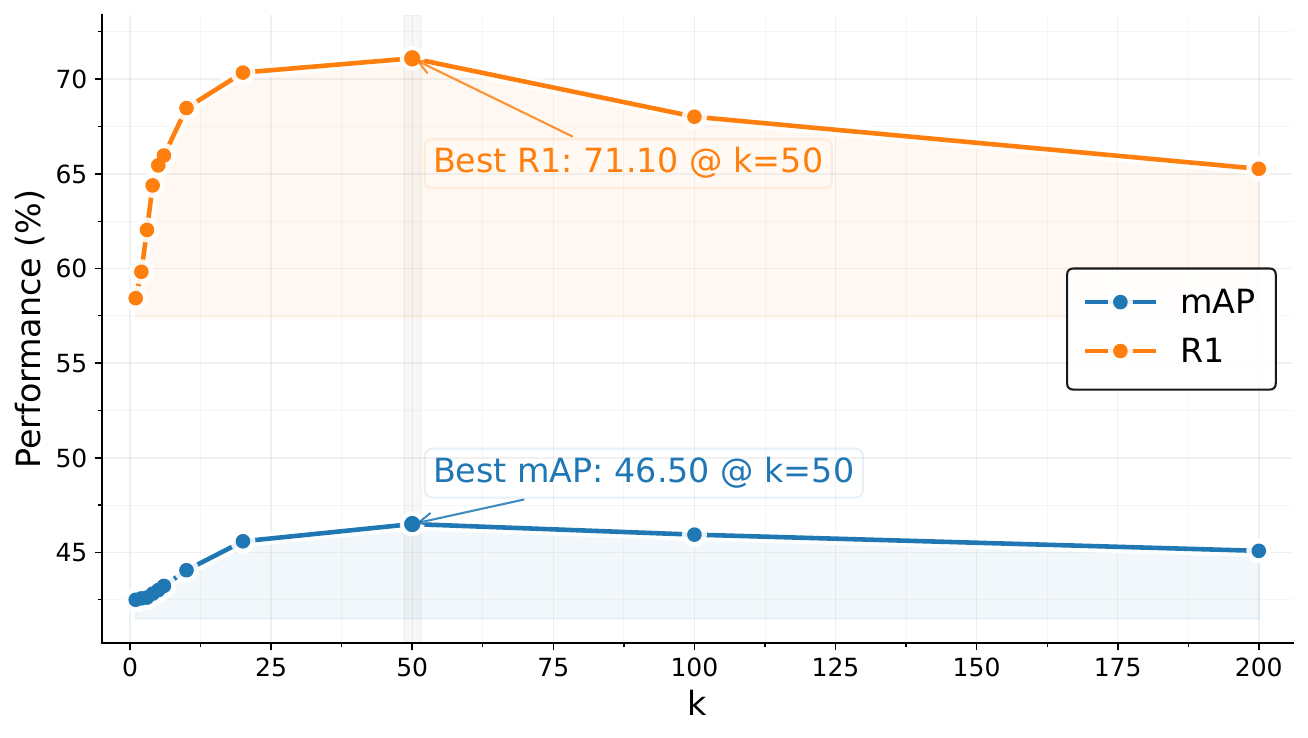}
        \caption{Top-k}
        \label{fig:ablation_topk}
    \end{subfigure}\hfill
    \begin{subfigure}[t]{0.18\linewidth}
        \centering
        \includegraphics[width=0.95\linewidth]{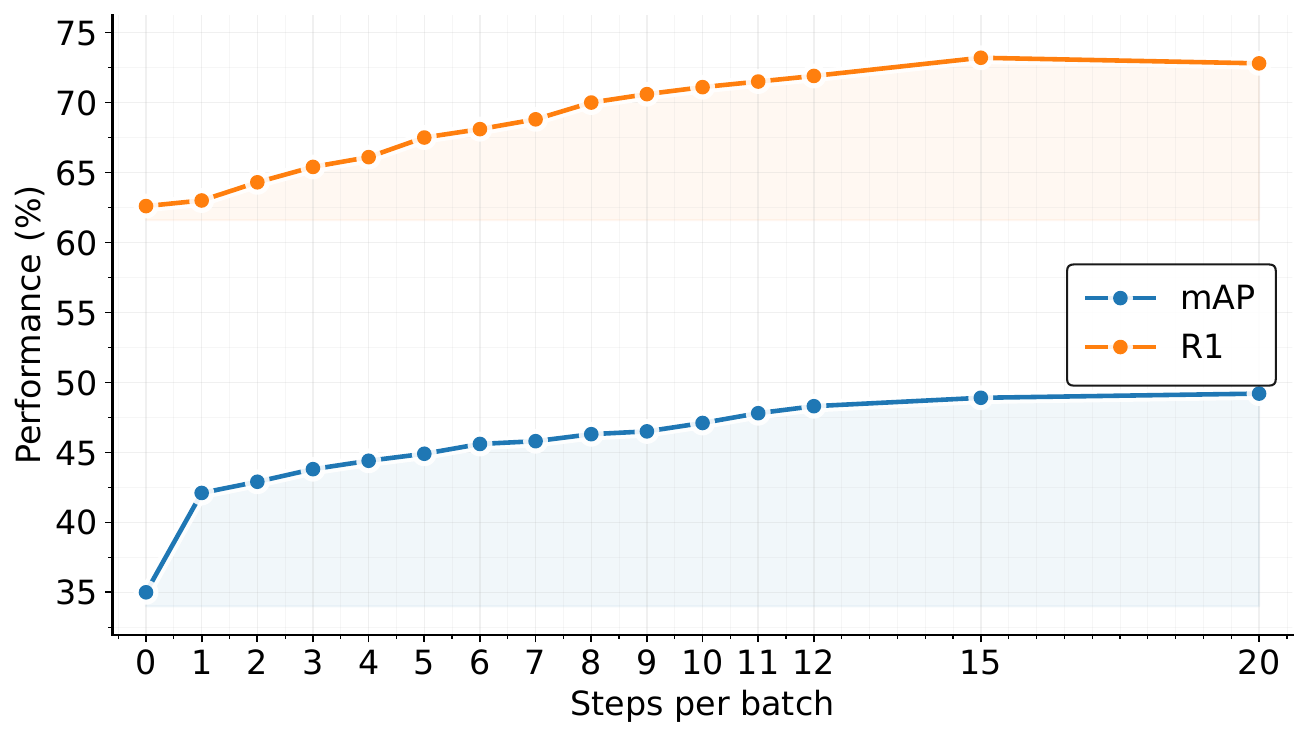}
        \caption{Steps per batch}
        \label{fig:steps}
    \end{subfigure}\hfill
    \begin{subfigure}[t]{0.18\linewidth}
        \centering
        \includegraphics[width=0.9\linewidth]{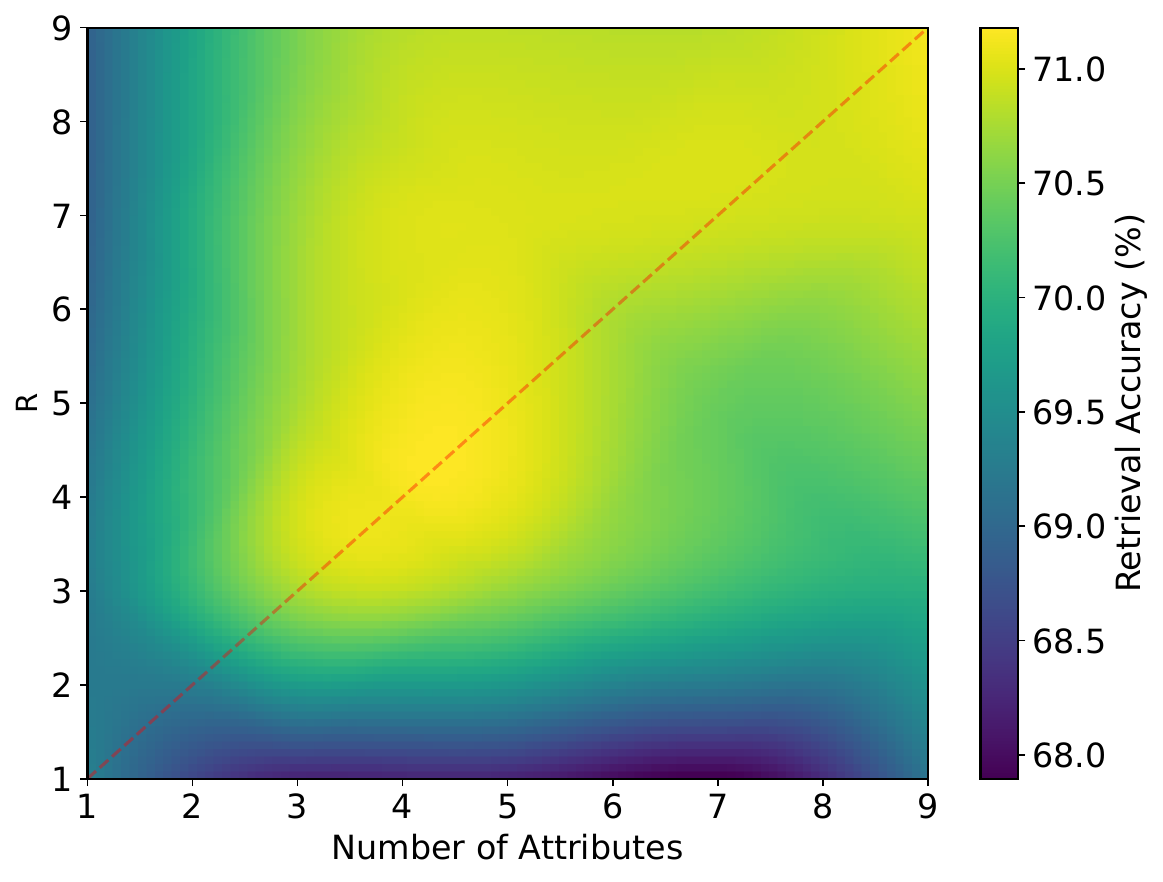}
        \caption{TopR vs number of attribute categories.}
        \label{fig:RvsattC}
    \end{subfigure}

    \vspace*{-6pt}

    \caption{\textbf{Hyperparameter Sensitivity.} Performance with respect to varying hyperparameters. Tested on Market-1501.}
    \label{fig:ablation_row}

    \vspace*{-12pt}
\end{figure*}

\vspace{-4.0pt}
\subsection{Hyperparameter Sensitivity}
\vspace{-4.0pt}
This section is aimed at measuring how sensitive the performance of \method{} is when the various hyperparameters are varied. While the sensitivity plots identify the best-performing values on Market-1501, all experiments in the manuscript were run using the same set of hyperparameters to show that \method{} is sufficiently robust to a hyperparameter set which is not explicitly optimized for a specific dataset.

\mypar{Effect of $\alpha_{\mathrm{acs}}$ (\autoref{fig:ablation_alpha}):}
$\alpha_{\mathrm{acs}}{=}0.1$ gives the best trade-off, achieving peak Rank-1 and near-peak mAP. Larger values gradually degrade performance, suggesting that overly strong ACS weighting can over-constrain the embedding with noisy or partial attribute cues and harm fine-grained identity separation.

\mypar{Effect of TopR (\autoref{fig:ablation_topr}):}
Performance peaks around $R{=}5$, which matches the number of main attribute categories used in our prompt bank. Smaller $R$ under-represents semantic  diversity, while larger $R$ introduces redundant or noisy directions that dilute the attribute-conditioned subspace.

\mypar{Effect of Top-k (\autoref{fig:ablation_topk}):}
Performance improves quickly from very small $k$ and then saturates, with the best Rank-1 around $k{=}50$. This reflects a selection trade-off: too small $k$ gives unstable pseudo-neighborhoods, whereas too large $k$ admits more false positives and weakens discrimination.

\mypar{Effect of optimization steps (\autoref{fig:steps}):}
Performance improves rapidly during the first test-time updates and then saturates after roughly $10$ steps. Since runtime grows nearly linearly with the number of steps, additional updates provide limited accuracy gains relative to their computational cost.

\mypar{Number of attributes vs TopR (\autoref{fig:RvsattC}):}
We find that accuracy generally improves with more attribute categories and moderately larger $\operatorname{TopR}$, but the interaction between the two is not strictly monotonic. This implies that diverse attribute categories provide complementary directions, while overly large $\operatorname{TopR}$ can add redundant prompts once sufficient semantic coverage is available.

\vspace{-3.0pt}
\subsection{Ablation Study}
\vspace{-2.0pt}

\autoref{tab:prompt_ablation} isolates the contributions of ACS, query conditioning, orthogonalization, score composition, and the adaptation objective. We also compare ACS with fixed low-rank PCA projections and stronger calibrated score-fusion baselines.

\noindent \textbf{Component Ablations.}
Using ACS only as a frozen score gives little improvement over the base model, while adaptation without ACS provides only modest gains. In contrast, combining query-side adaptation with ACS improves Market-1501 from 35.0/62.6 to 46.5/71.1 mAP/R1 and RSTPReID from 23.1/34.5 to 27.6/37.6. Thus, ACS is not effective as a standalone re-ranking signal; its main benefit arises from guiding the adaptation process with query-conditioned semantic geometry.

\mypar{Subspace and Score-Fusion Ablations.}
Shuffling prompts, using a constant ACS, or replacing ACS with a random subspace substantially degrades performance, confirming the importance of both semantic grounding and query conditioning. Removing QR while keeping the same selected directions also reduces performance to 44.2/68.1 on Market-1501 and 26.4/36.8 on RSTPReID, showing that orthogonalization is beneficial. Motivated by a denoising interpretation, we further compare matched-rank PCA subspaces from the prompt bank and gallery features. Both are competitive in mAP, reaching 45.3/63.0 and 45.2/62.9 on Market-1501, but remain below ACS, especially in Rank-1 accuracy, with the same ordering on RSTPReID. Thus, low-rank projection itself is useful, while query-conditioned semantic selection better resolves top-ranked matches. Stronger score-fusion baselines show a similar trend. Standardized and temperature-calibrated fusion improve over direct fusion, but the best variant reaches only 41.5/61.8 and 25.1/33.9, remaining clearly below \method{}.

\vspace{-3.0pt}
\begin{table}[!h]
\centering
\caption{
\textbf{Ablations.}
All rows use the same TopR, rank $r$, number of adaptation steps, and $\alpha_{\mathrm{acs}}$, unless stated otherwise. I2I-ReID results are reported on Market-1501 and T2I-ReID results on RSTPReID.
}
\label{tab:prompt_ablation}

\vspace*{-5pt}

\resizebox{0.8\linewidth}{!}{
\setlength{\tabcolsep}{5pt}
\begin{tabular}{llcccc}
\toprule
\multirow[b]{2}{*}{\textbf{Group}} & \multirow[b]{2}{*}{\textbf{Variant}} 
& \multicolumn{2}{c}{\textbf{I2I-ReID}} 
& \multicolumn{2}{c}{\textbf{T2I-ReID}} \\
\cmidrule(lr){3-4}\cmidrule(lr){5-6}
& & mAP & R1 & mAP & R1 \\
\midrule

\multirow{4}{*}{\makecell[l]{Component \\ Ablation}}
& No adapt 
& 35.0 & 62.6 & 23.1 & 34.5 \\
& ACS only (No-adaptation, $\alpha_{\mathrm{acs}}{=}0.1$)
& 35.2 & 61.8 & 23.6 & 34.1 \\
& Adaptation w/o ACS $(\alpha_{\mathrm{acs}}{=}0)$
& 36.8 & 65.0 & 23.5 & 35.4 \\
& ACS adaptation (i.e. \method{}) 
& \textbf{46.5} & \textbf{71.1} & \textbf{27.6} & \textbf{37.6} \\

\midrule
\multirow{6}{*}{\makecell[l]{Subspace \\ Ablation}}
& Shuffled prompts across queries 
& 37.8 & 62.8 & 23.5 & 35.3 \\
& Constant ACS subspace 
& 36.7 & 62.6 & 23.5 & 35.1 \\
& Random subspace 
& 35.1 & 62.9 & 24.2 & 34.8 \\
& ACS w/o QR orthogonalization
& 44.2 & 68.1 & 26.4 & 36.8 \\
& Global prompt PCA (rank $r{=}5$)
& 45.3 & 63.0 & 27.0 & 36.9 \\
& Gallery feature PCA (rank $r{=}5$)
& 45.2 & 62.9 & 26.8 & 36.8 \\

\midrule
\multirow{3}{*}{
\makecell[l]{Score Fusion \\Control}
}
& Direct score fusion
& 39.2 & 60.3 & 24.5 & 32.9 \\
& Standardized ($z$-score) fusion
& 41.4 & 61.1 & 24.8 & 33.1 \\
& Temperature-calibrated probability fusion
& 41.5 & 61.8 & 25.1 & 33.9 \\

\midrule
\multirow{5}{*}{
\makecell[l]{Adaptation \\Objective}
}
& Top-1 pseudo-label NLL
& 42.7 & 58.2 & 24.8 & 34.6 \\
& Full retrieval entropy (no top-$k$)
& 41.1 & 61.0 & 25.1 & 34.7 \\
& Top-$k$ entropy
& 45.3 & 63.1 & 26.8 & 35.9 \\
& Raw top-$k$ similarity (w/o softmax)
& \phz6.2 & 11.5 & \phz4.1 & \phz9.3 \\
& Proposed top-$k$ set likelihood (Eq.~7)
& \textbf{46.5} & \textbf{71.1} & \textbf{27.6} & \textbf{37.6} \\

\midrule
\multirow{4}{*}{
\makecell[l]{Template Bias \\Control}
}
& Template Corruption, Attributes Preserved
& 46.1 & 69.7 & 27.3 & 37.2 \\
& Attribute tokens only 
& 44.2 & 69.1 & 26.7 & 37.8 \\
& w/o Subtraction 
& 40.7 & 62.8 & 24.3 & 36.6 \\
& w/o Subtraction + Template Corruption 
& 40.5 & 62.7 & 24.4 & 36.6 \\
\bottomrule
\end{tabular}
}

\vspace*{-10pt}
\end{table}

\mypar{Adaptation Objective.}
We compare Eq.~(7) with several alternatives, including top-$k$ entropy minimization\footnote{Top-$k$ entropy replaces $\log \mathbf{p}_i(j)$ in Eq.~(7) with $\mathbf{p}_i(j)\log \mathbf{p}_i(j)$.}. Top-1 NLL and full-gallery entropy reach 42.7/58.2 and 41.1/61.0 on Market-1501, while top-$k$ entropy improves to 45.3/63.1. The proposed set likelihood remains best at 46.5/71.1, with the same trend on RSTPReID (27.6/37.6 versus 26.8/35.9). Unlike top-$k$ entropy, Eq.~(7) promotes the selected neighborhood collectively rather than explicitly sharpening probabilities within it, avoiding premature commitment to individual candidates. Raw top-$k$ similarity without softmax performs poorly, showing the importance of gallery-relative normalization.

\mypar{Template Bias Control.}
Corrupting the shared template while preserving attribute content leaves performance nearly unchanged, showing limited sensitivity to exact phrasing. Attribute tokens alone cause a moderate drop, while removing neutral-feature subtraction reduces Market-1501 mAP from 46.5 to 40.7. Additional template corruption without subtraction changes little, indicating that centering is the main mechanism suppressing shared template bias.


\vspace{-6.0pt}
\section{Conclusion}
\label{sec:concl}
\vspace{-4.0pt}

\noindent\textbf{Limitations. } \method{} uses a fixed standardized attribute bank, so its effectiveness depends on the coverage of the prompt vocabulary and may be limited for rare or highly fine-grained appearance cues. Furthermore, when the initial VLM representation is very weak or the query is severely occluded/noisy, the selected attribute directions may be unreliable, limiting the benefit of ACS. This has been analyzed and reported in the Supplementary Materials.

We introduced \method{}, a label-free TTA framework for VLM-based ReID across I2I and T2I retrieval. Our results show that attribute prompts are unreliable as direct matching scores, but become effective when used as \emph{query-conditioned directions} for guiding retrieval adaptation. This provides a lightweight way to inject attribute consistency at test time without identity or attribute labels. Future work will explore query-adaptive prompt banks to improve coverage beyond a fixed bank of pedestrian attributes.

{
    \small
    \bibliographystyle{ieeenat_fullname}
    \bibliography{references}
}
\clearpage
\clearpage
\appendix


\clearpage

\twocolumn[
\begin{center}
{\Large \bfseries Supplementary Material}

\vspace{1em}

{\large In \method{} of Attributes: Grounding Retrieval Geometry with Query-Conditioned Subspaces for Person Re-Identification}

{Code and Attribute Bank: \href{https://github.com/rB080/praise.git}{ \faGithub Github}}
\end{center}

\vspace{2em}
]
\maketitle


\section{Geometric Motivation for Attribute-Conditioned Retrieval}
\label{sec:method:theory}

\mypar{Prompt directions as biased semantic displacements.}
The central role of ACS is to use attribute prompts as directions that shape the retrieval geometry, rather than as calibrated attribute classifiers. To motivate this design, consider a normalized prompt embedding $\zz_{t,\ell}=\tenc(s_\ell)$. Because all prompts share a similar person-centric template, their embeddings contain both an attribute-specific component and a common template component. We model this as
\begin{equation}
    \zz_{t,\ell}
    =
    \mathbf{b}
    +
    \mathbf{a}_{\ell}
    +
    \boldsymbol{\epsilon}_{\ell},
    \label{eq:prompt_decomposition}
\end{equation}
where $\mathbf{b}$ denotes the shared person/template component,
$\mathbf{a}_{\ell}$ denotes the semantic displacement induced by the attribute suffix, and $\boldsymbol{\epsilon}_{\ell}$ captures lexical and encoder noise. For a query embedding $\zz_i$, direct prompt activation gives:
\begin{equation}
    \zz_i^\top \zz_{t,\ell}
    =
    \zz_i^\top \mathbf{b}
    +
    \zz_i^\top \mathbf{a}_{\ell}
    +
    \zz_i^\top \boldsymbol{\epsilon}_{\ell}.
\end{equation}
The first term is shared across many prompts and can dominate the
attribute-specific term. Thus, raw prompt responses may be strongly affected by template bias and should not be treated as calibrated matching scores.

To reduce this shared component, \method{} subtracts the neutral prompt embedding $\zz_{t,0}=\tenc(s_0)$ and uses centered directions
\begin{equation}
    \bar{\zz}_{t,\ell}
    =
    \zz_{t,\ell}-\zz_{t,0}.
    \label{eq:theory_centering}
\end{equation}
If the neutral prompt captures the common person/template component, i.e.,
$\zz_{t,0}\approx \mathbf{b}$, then
\begin{equation}
    \bar{\zz}_{t,\ell}
    \approx
    \mathbf{a}_{\ell}
    +
    \boldsymbol{\epsilon}_{\ell}
    -
    \boldsymbol{\epsilon}_{0}.
\end{equation}
Therefore, prompt centering can be interpreted as a first-order removal of template bias, leaving a direction that better approximates the attribute-specific
semantic displacement.

\mypar{Why direct attribute-profile fusion is unstable.}
Let the query and gallery prompt-response profiles be $\mathbf{u}_i=\bar{\ZZ}_t\zz_i^\top$ and $\mathbf{v}_j=\bar{\ZZ}_t\zz_{g,j}^\top$. Ignoring the final normalization in \autoref{eq:bg_direct_attr_score}, their inner product can be written as
\begin{equation}
    \mathbf{u}_i^\top \mathbf{v}_j
    =
    \zz_i^\top
    \bar{\ZZ}_t^\top \bar{\ZZ}_t
    \zz_{g,j}.
    \label{eq:direct_profile_metric}
\end{equation}
Thus, direct attribute-profile similarity is equivalent to comparing query and gallery embeddings under the fixed metric
$\mathbf{M}_t=\bar{\ZZ}_t^\top\bar{\ZZ}_t$. This metric depends on the full prompt bank, including redundant and correlated prompts. If many prompts encode nearby concepts, such as similar colors or clothing phrases, then $\mathbf{M}_t$ over-weights those correlated directions. Consequently, $\attmx^{\text{direct}}$ can amplify coarse attribute overlap while failing to separate true identity matches from distractors that share common attributes. This motivates replacing global prompt-profile fusion with a query-conditioned subspace comparison.

\mypar{ACS as a query-conditioned semantic projector.}
For each query $\zz_i$, \method{} selects the most compatible centered prompt directions,
\begin{equation}
    \bA_i=\operatorname{TopR}(\bar{\ZZ}_t\zz_i^\top),
    \qquad |\bA_i|=r.
\end{equation}
This estimates a small set of active person-centric factors for the query. Since only a few attributes are usually relevant for a particular person image or text description, using the full prompt bank may introduce irrelevant directions. In contrast, $\bA_i$ defines a local semantic coordinate system specific to query $i$.

Let
\begin{equation}
    \mathbf{U}_i
    =
    \left[
    \bar{\zz}_{t,\ell_1}
    \cdots
    \bar{\zz}_{t,\ell_r}
    \right]^\top
    \in\Real^{r\times d},
    \qquad \ell_k\in\bA_i.
\end{equation}
The ACS basis is then obtained by thin QR decomposition:
\begin{equation}
    (\mathbf{U}_i^{w})^\top
    =
    \QQ_i\mathbf{R}_i,
    \qquad
    \QQ_i^\top\QQ_i=I_m,
    \qquad m\le r.
\end{equation}

The important object is the projector
\begin{equation}
    \PPi_i^\parallel=\QQ_i\QQ_i^\top.
\end{equation}
QR does not introduce new semantic information; it only constructs an orthonormal basis for the span of the selected attribute directions. Hence, ACS depends on the selected semantic span, not on the arbitrary scale,
ordering, or redundancy of the prompts. In particular, for any orthogonal matrix $\mathbf{O}\in\Real^{m\times m}$,
\begin{equation}
    (\QQ_i\mathbf{O})(\QQ_i\mathbf{O})^\top
    =
    \QQ_i\QQ_i^\top
    =
    \PPi_i^\parallel.
\end{equation}
Therefore, the induced projection is invariant to the particular orthonormal basis returned by QR.

\mypar{Projected similarity as attribute-consistent evidence.}
The ACS affinity compares query and gallery features only through coordinates inside the query-conditioned semantic subspace:
\begin{equation}
    \bAA^\parallel[i,j]
    =
    (\tilde{\zz}_i^\parallel)^\top
    \tilde{\zz}_{g,j}^\parallel,
    \qquad
    \zz_i^\parallel=\PPi_i^\parallel\zz_i,
    \quad
    \zz_{g,j}^\parallel=\PPi_i^\parallel\zz_{g,j}.
\end{equation}
Equivalently, since $\QQ_i^\top\QQ_i=I_m$, this can be computed using subspace coordinates
\begin{equation}
    \mathbf{c}_i=\zz_i^\top\QQ_i,
    \qquad
    \mathbf{c}_{g,j}=\zz_{g,j}^\top\QQ_i,
\end{equation}
followed by $\ell_2$ normalization and an inner product. This shows that ACS is not a prompt-response matching score. Instead, it is a projected cosine similarity under a query-conditioned semantic projector.

This distinction is important for ReID. Attributes such as clothing color, gender, or accessories are not identity labels; many identities may share them. Therefore, $\bAA^\parallel$ should not replace the identity affinity $\idmx$. Instead, it provides auxiliary evidence that checks whether a candidate gallery image agrees with the query along the selected semantic factors. The final affinity is therefore
\begin{equation}
    \bAA[i,j]
    =
    \idmx[i,j]
    +
    \alpha_{\mathrm{acs}}\bAA^\parallel[i,j].
    \label{eq:theory_final_affinity}
\end{equation}
A small $\alpha_{\mathrm{acs}}$ makes ACS act as a semantic regularizer on the identity geometry: candidates must remain close in the original embedding space, but are additionally favored when they agree along query-relevant attribute directions.

\section{Motivation for the Loss function}
\label{sec:supp:loss_motivation}

\begin{figure}[!ht]
    \centering
    \begin{subfigure}[t]{0.48\columnwidth}
        \centering
        \includegraphics[width=\linewidth]{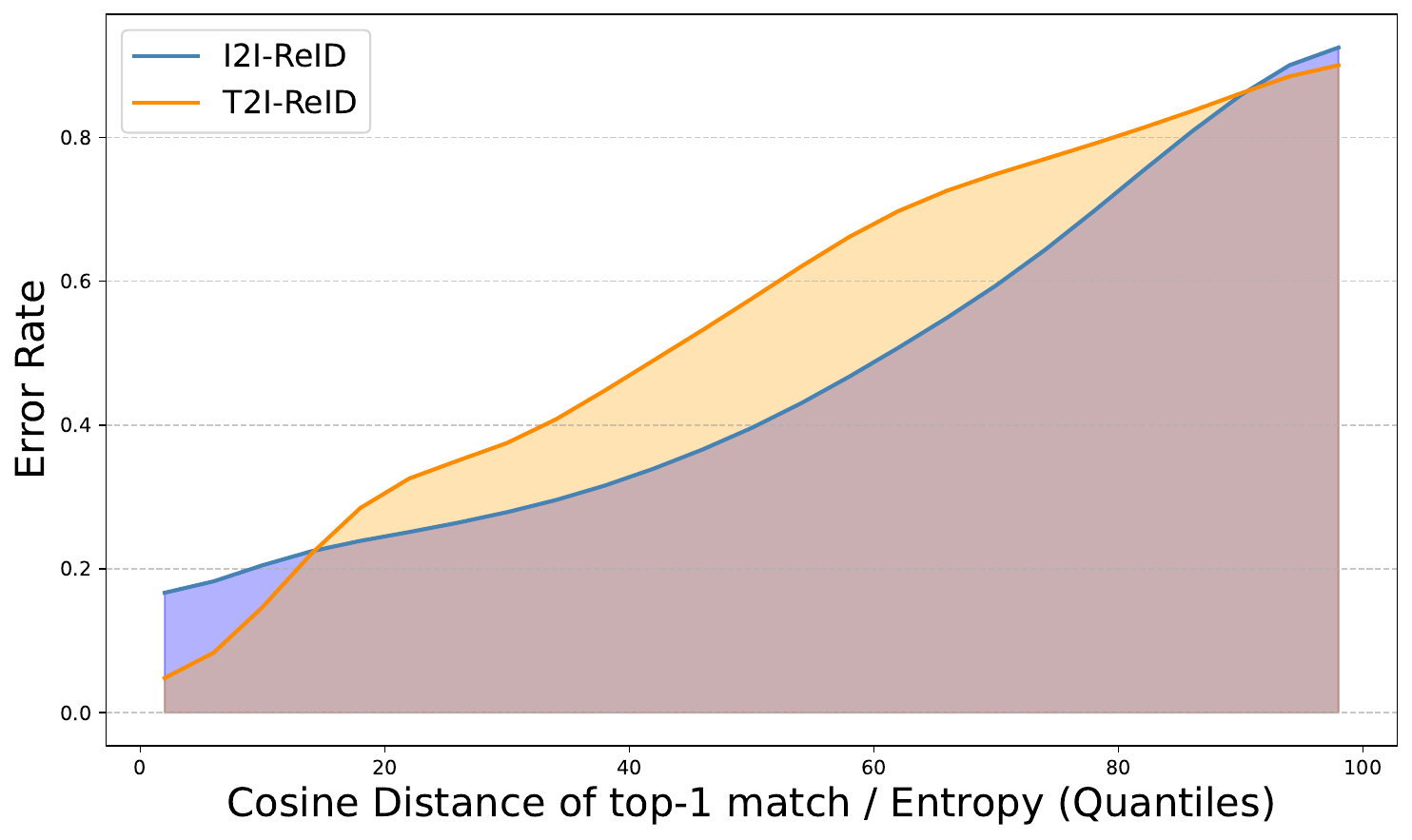}
        \caption{Error rate vs Top-1 Cosine Distance}
        \label{fig:edist}
    \end{subfigure}\hfill
    \begin{subfigure}[t]{0.48\columnwidth}
        \centering
        \includegraphics[width=\linewidth]{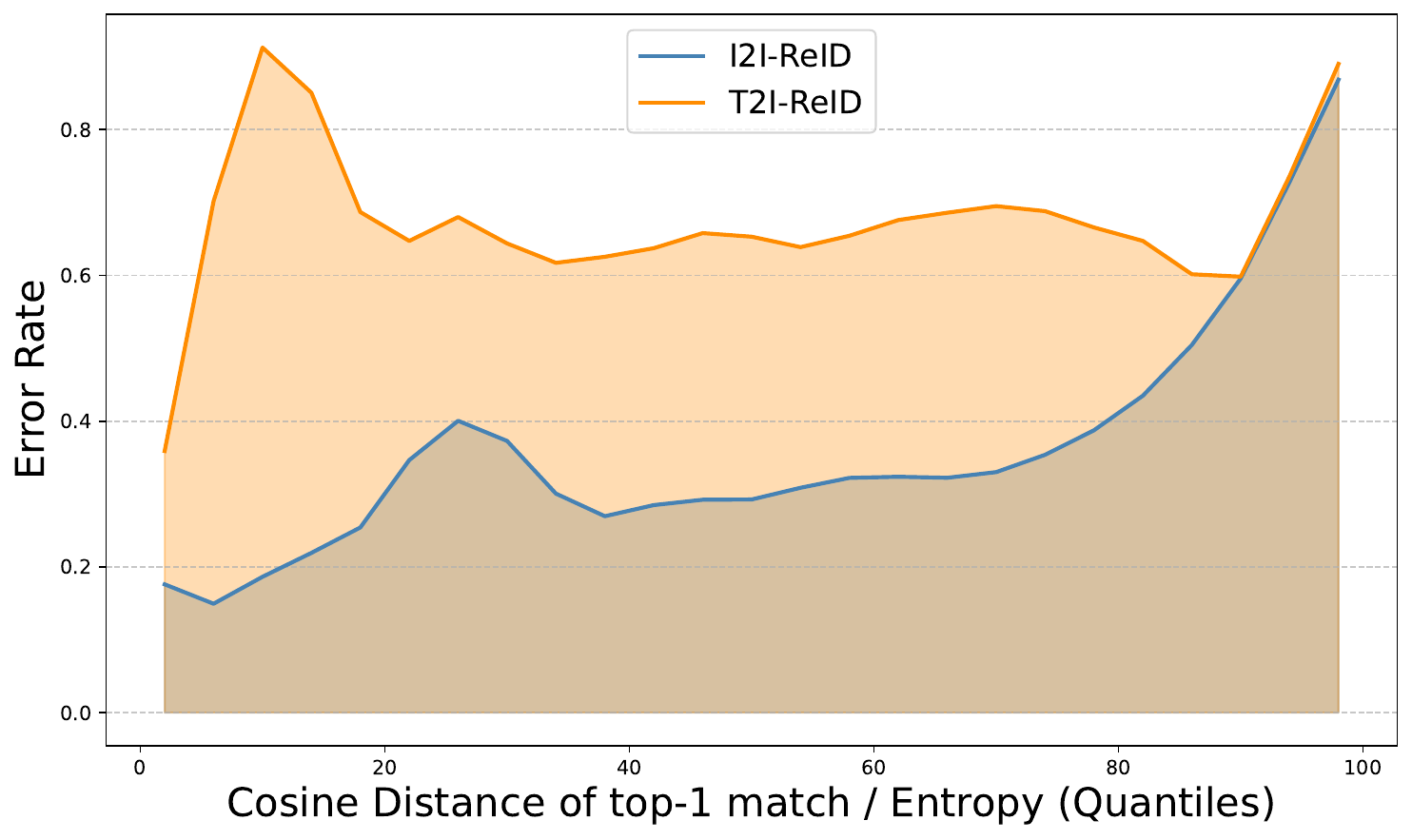}
        \caption{Error rate vs Feature Entropy}
        \label{fig:eent}
    \end{subfigure}

    \caption{Retrieval Top-1 Error-rate distribution across entropy and cosine distance. Curves computed on the Market-1501 \cite{zheng2015scalable} dataset for I2I-ReID and RSTPReID \cite{zhu2021dssl} for T2I-ReID.}
    \label{fig:ent_vs_dist}

    \vspace*{-10pt}
\end{figure}

We begin this section with a key observation in our loss function (Eq. 7, Main manuscript), that the loss essentially is a \emph{minimization of cosine distance} within top-k gallery neighbours. Critically, most common TTA methods employ a direct feature level entropy \cite{wang2020tent, niu2022efficient, shu2022test}. We first consider an empirical analysis to observe trends in retrieval error rate versus (1) cosine distance and (2) feature entropy. Thereafter, we use these findings to develop a theoretical explanation for why direct entropy is \emph{ill-posed} as a potential retrieval uncertainty surrogate, and an alternative loss function that is suitable for ReID.

\subsection{Empirical evidence: why feature-level entropy is a weak proxy for ReID uncertainty}
\label{sec:supp:entropy_pitfall}

We report two diagnostic curves (\autoref{fig:ent_vs_dist}) that relate the \emph{test error rate} to two candidate ``confidence'' signals computed from the frozen retrieval model: (i) the \emph{cosine distance of the top-1 match between query and gallery}, and (ii) the \emph{feature level entropy} of a prediction distribution. The distance plot exhibits a clear, smooth, and largely monotonic trend: as the top-1 cosine distance increases, the error rate increases substantially. This behavior is consistent with the ReID geometry, i.e. when the closest gallery point is far in the embedding space, the query is less likely to have a correct nearest neighbor. In contrast, the entropy plot shows a noisier and partially non-monotonic relationship. In particular, there exist regimes where entropy is relatively low but the error rate is high, and also regimes where entropy grows without a predictable change in error. 
This implies that direct feature level entropy is \emph{miscalibrated} as an uncertainty indicator for ReID.

Entropy minimization implicitly encourages \emph{peaky} predictions. In ReID test-time evaluation, identities are disjoint from training, and the correct notion of confidence is whether the query embedding is near its \emph{true} neighbor(s) in $\ZZ_g$, not whether it yields a confident classification under some surrogate label space. Concretely, entropy is typically defined over a distribution formed with respect to a classifier head trained on source identities. In case of ReID/Retrieval, the peaky predictions can occur for the \emph{wrong reasons}. For example, a model can become highly confident in a direction that is unrelated to the correct gallery neighborhood, yielding low entropy but high error. This matches the empirical curve where low entropy does not guarantee low error.

Therefore, directly minimizing feature-level entropy is risky in ReID because it can over-emphasize ``peakedness'' (confidence) rather than \emph{neighborhood correctness}. Under domain shift, this may accelerate drift toward spurious modes (overconfident wrong matches), which is exactly the kind of collapse we aim to avoid in test-time adaptation.

\subsection{Theoretical justification: distance as an uncertainty surrogate for nearest-neighbor retrieval}
\label{sec:supp:theory_distance}

We provide an intuition consistent with the observed monotonicity in the distance plot.

\begin{proposition}[Distance as an uncertainty surrogate]
\label{prop:dist_uncertainty}
Let $\zz_i$ be a query feature and $\zz_{g,j}$ be a gallery feature in $\Real^d$.\footnote{For simplicity, we drop the tilde ($\, \tilde{} \,$) sign as all features discussed \emph{in this section} is assumed to be normalized.} Consider a nonnegative function $u(\cdot)$ that models uncertainty as a function of the Cosine distance between features, i.e.,
\[
u(d)\quad \text{with} \quad d \;=\; 1 - \zz_i^\top \zz_{g,j}.
\]
A natural uncertainty surrogate satisfies:
(i) $u(0)=0$ (certainty at zero distance) and
(ii) $u$ is non-decreasing (larger distance $\Rightarrow$ larger uncertainty).
\end{proposition}

\mypar{Interpretation.}
When the closest gallery neighbor to $\zz_i$ lies far away, there is more room for an incorrect identity to become the nearest neighbor under small perturbations (domain shift, camera style changes, background clutter). Hence, larger distance should correspond to higher retrieval uncertainty, consistent with the monotonic trend in the cosine-distance plot.

\begin{proposition}[Nearest-neighbor error is non-decreasing with distance]
\label{prop:nn_error}
Fix a gallery set $\{\zz_{g,j}\}_{j=1}^{N_g}\subset \Real^d$ where $\zz_{g,1}$ denotes the unique correct gallery embedding for query $\zz_i$ and all other points are incorrect. Consider the 1-NN decision rule $\hat{\jmath}(\zz_i)=\arg\min_j(1 - \zz_i^\top \zz_{g,j})$. If the conditional distribution of $\zz_i$ given $1 - \zz_i^\top \zz_{g,j}=r$ is isotropic around $\zz_{g,1}$, then the probability of error
\[
e(r)=\Pr\!\big[\hat{\jmath}(\zz_i)\neq 1\;\big|\;\|\zz_i-\zz_{g,1}\|_2=r\big]
\]
is a non-decreasing function of $r$.
\end{proposition}

\begin{figure}[t]
    \centering
    \begin{subfigure}[t]{0.43\columnwidth}
        \centering
        \includegraphics[width=\linewidth]{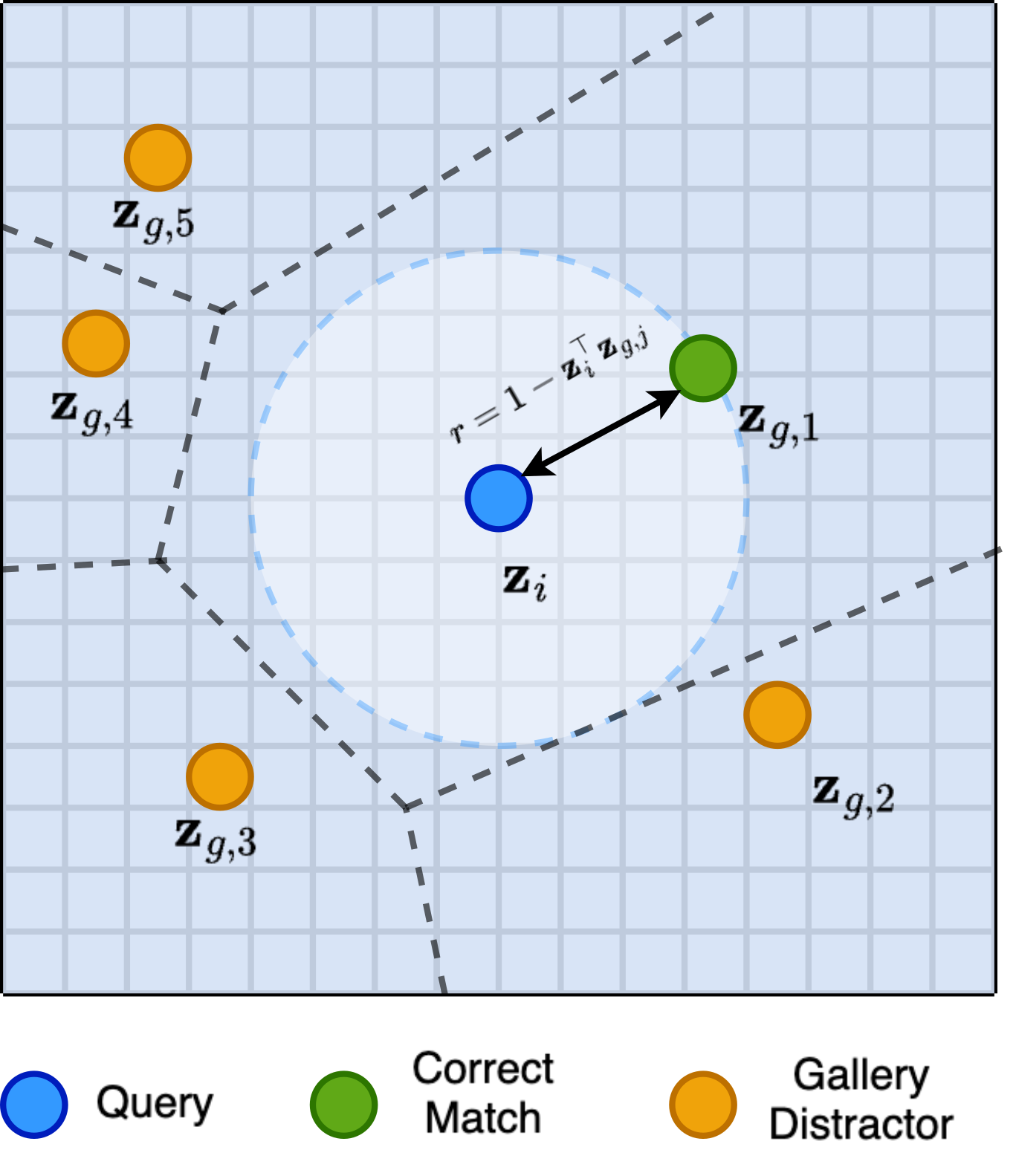}
        \caption{Query-Gallery Embedding Geometry}
        \label{fig:emb_geo}
    \end{subfigure}\hfill
    \begin{subfigure}[t]{0.42\columnwidth}
        \centering
        \includegraphics[width=\linewidth]{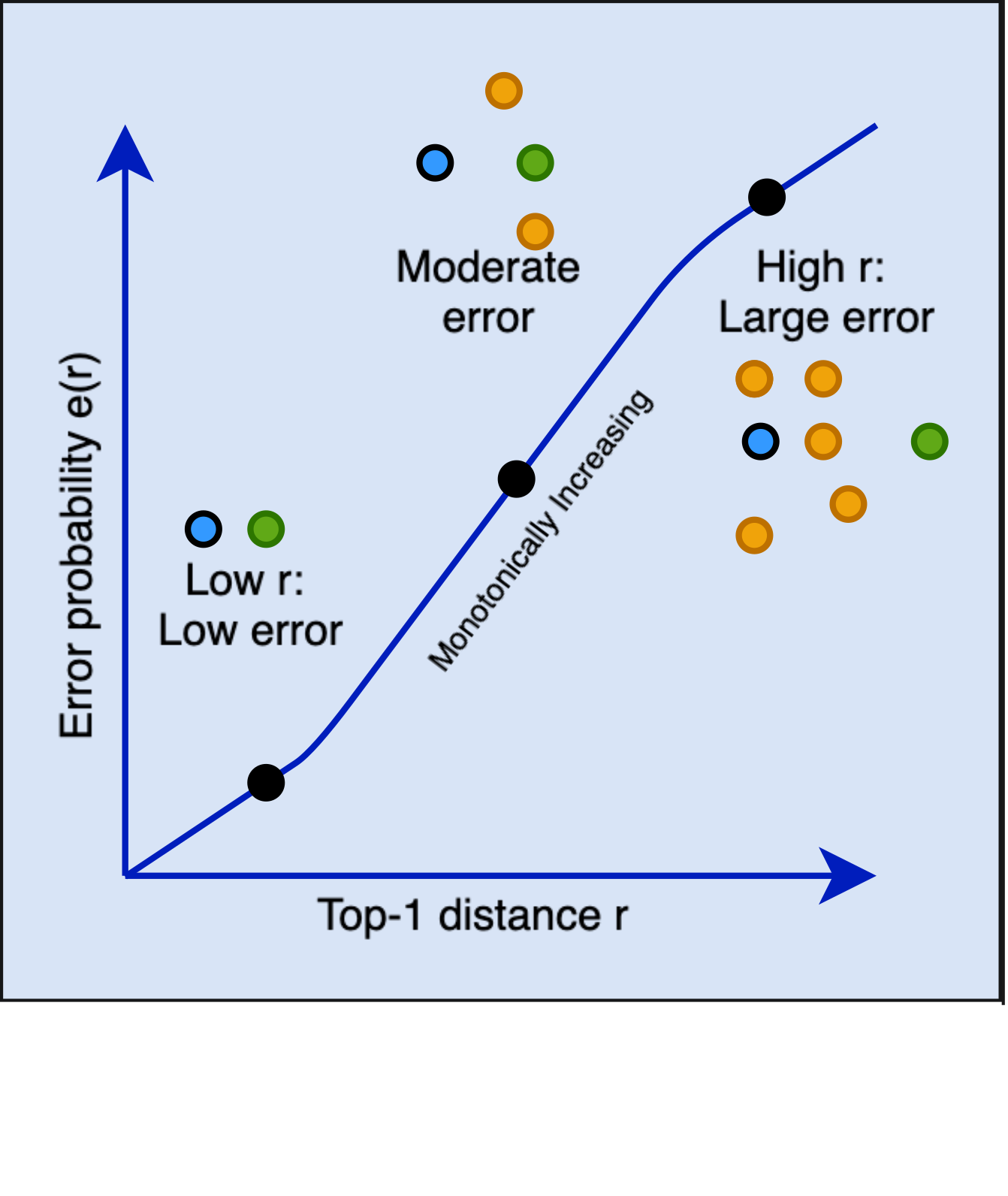}
        \caption{$e(r)$ vs $r$ expected curve}
        \label{fig:er_r}
    \end{subfigure}

    \caption{Understanding the nature of the embedding space between a query and gallery, and a theoretical interpretation of \autoref{fig:ent_vs_dist} where error rate is a surrogate for $e(r)$.}
    \label{fig:embedding_space}

    \vspace*{-10pt}
\end{figure}

\mypar{Discussion.}
As illustrated in \autoref{fig:embedding_space}, for fixed gallery points, the set of query locations that are correctly classified by 1-NN forms a Voronoi region around $\zz_{g,1}$. Intersecting this region with a sphere of radius $r$ centered at $\zz_{g,1}$ yields the set of directions from which the correct point remains nearest. As $r$ increases, this intersection cannot gain new directions, and thus the measure of ``correct directions'' does not increase. Under isotropy, this implies $e(r)$ is non-decreasing. This formalizes the key empirical observation: \emph{larger nearest-neighbor distance implies a higher chance of top-1 retrieval error.}

Our ACS affinity $\bAA[i,j]$ in \eqref{eq:final_acs_affinity} is still a geometric similarity between query and gallery features (augmented by the ACS-induced subspace alignment). Thus, sharpening retrieval under $\bAA$ is aligned with increasing confidence in a \emph{distance-consistent} way, unlike feature-level entropy that may be defined over a mismatched label space.

\subsection{Pitfalls of using cosine distance directly as loss}
\label{sec:supp:why_softmax_topk}

A naive objective would be to reduce the distance to the current best match, e.g.,
\[
\min_\theta \;\frac{1}{B}\sum_{i=1}^B \big(1-\bAA[i,\hat{\jmath}_i]\big),
\qquad \hat{\jmath}_i=\arg\max_j \bAA[i,j].
\]
However, this is unstable for large-scale retrieval: (1) it depends on the discrete $\arg\max$ index $\hat{\jmath}_i$, which can switch abruptly across steps, producing high-variance gradients; (2) it over-commits to a single neighbor early, making the method vulnerable to confirmation bias under domain shift (a wrong early top-1 can be reinforced aggressively); and (3) its gradients do not explicitly control the \emph{relative} separation between a small candidate set and the rest of the gallery.

Instead, we define the retrieval distribution: $\mathbf{p}_i = \operatorname{softmax}(\bAA[i,:]/\tau),$ which replaces hard assignment with a differentiable relaxation. The exponential map amplifies meaningful logit gaps while remaining smooth, and the temperature $\tau$ explicitly controls sharpness, i.e. high $\tau$ yields conservative updates and lower $\tau$ yields stronger sharpening. This improves optimization stability and makes the loss less sensitive to small perturbations in $\bAA$.

Using $-\log \mathbf{p}_i(j)$ yields well-behaved gradients as it focuses optimization on increasing probability mass rather than unboundedly increasing raw similarities, and it naturally operates on \emph{relative} scores across the gallery. In practice, this reduces the tendency to exploit trivial scaling effects and provides consistent gradients even when similarities saturate.

A full entropy minimization over all $N_g$ gallery entries encourages the entire distribution to become extremely peaky, which can collapse to an arbitrary wrong index when the model is miscalibrated (as suggested by the entropy plot). We therefore minimize only over the head:

\begin{equation} \label{supp:eq:loss}
    \mathcal{L}_{\text{TTA}}
    \,=\,
    -\frac{1}{B}\sum_{i=1}^{B}\sum_{j\in\mathcal{K}_i}\log \mathbf{p}_i(j) \qquad \mathcal{K}_i=\operatorname{TopK}(\mathbf{p}_i).
\end{equation}

This has three stabilizing effects: (i) it avoids over-penalizing the long tail of the gallery, which is dominated by many near-zero probabilities and can inject large aggregate gradient noise when $N_g$ is large; (ii) it reduces confirmation bias by allowing a \emph{small candidate set} to remain plausible rather than forcing a single winner too early; and (iii) it mitigates catastrophic collapse: even if the current top-1 is wrong, concentrating probability mass over several top candidates is empirically more robust and permits recovery across adaptation steps.

\section{Differentiability of the Loss}
\label{supp:diff}

\method{} frequently interleaves discrete retrieval operators (Top-$k$/Top-$r$) with continuous projection and normalization steps. As a result, it is important to assess differentiability to make explicit which components support gradient-based test-time optimization and which are treated as stop-gradient index selections.

\mypar{Setup.}
Let $\theta$ denote the adapted parameters (the affine scale and bias in normalization layers) in the encoder being adapted. For a query batch $\XX_q$ of size $B$, the query embeddings are
\[
\ZZ(\theta)=\venc(\XX_q;\theta)\in\Real^{B\times d},
\qquad \zz_i(\theta)\in\Real^d \text{ is row } i.
\]
We treat cached gallery embeddings $\ZZ_g\in\Real^{N_g\times d}$ as constants during TTA (i.e., $\partial \ZZ_g/\partial \theta = 0$), consistent with the implementation where $\ZZ_g$ is precomputed once per test set.

Given logits $\bAA(\theta)\in\Real^{B\times N_g}$, the gradient follows by the chain rule:
\[
\nabla_\theta \mathcal{L}
=
\sum_{i=1}^{B}
\left(\frac{\partial \zz_i}{\partial \theta}\right)^\top
\left(\frac{\partial \mathcal{L}}{\partial \zz_i}\right).
\]
Thus it suffices to characterize $\partial \mathcal{L}/\partial \bAA$ and $\partial \bAA/\partial \zz_i$.

\subsection{Differentiability of the retrieval loss}
\label{supp:diff_loss}

For each query $i$, define logits $\mathbf{s}_i=\bAA[i,:]\in\Real^{N_g}$ and retrieval distribution
\[
\mathbf{p}_i=\operatorname{softmax}(\mathbf{s}_i/\tau),
\qquad
p_i(j)=\frac{\exp(s_{i,j}/\tau)}{\sum_{m=1}^{N_g}\exp(s_{i,m}/\tau)}.
\]
The loss function has been defined in \autoref{supp:eq:loss}. This objective is \emph{piecewise smooth} in $\mathbf{s}_i$: it is smooth whenever the set $\mathcal{K}_i$ is unchanged (i.e., away from boundary events / ties). In implementation, $\mathcal{K}_i$ is treated as an index set (stop-gradient through $\operatorname{TopK}$).

Now, let $m_i(\ell)=\mathbb{I}[\ell\in\mathcal{K}_i]$. For a fixed $\mathcal{K}_i$,
\[
\frac{\partial \mathcal{L}_{\text{TTA}}}{\partial s_{i,\ell}}
=
\frac{1}{B\tau}\left(k\,p_{i,\ell}-m_i(\ell)\right),
\qquad \ell\in\{1,\ldots,N_g\}.
\]
This follows from $\partial \log p_{i,j}/\partial s_{i,\ell}=\tfrac{1}{\tau}(\mathbb{I}[\ell=j]-p_{i,\ell})$ (derivative of softmax function) and summing over $j\in\mathcal{K}_i$.

At ties around the $k$-th threshold, $\operatorname{TopK}$ is not differentiable. The loss remains well-defined, and the above expression corresponds to a valid subgradient under a deterministic tie-breaking rule.

\subsection{ACS affinity as a differentiable map}
\label{supp:diff_acs}

\mypar{Recap}
Given centered prompt directions $\bar{\ZZ}_t\in\Real^{L\times d}$ (constant) and a query embedding $\zz_i$,
prompt activations are $\mathbf{u}_i=\bar{\ZZ}_t\,\zz_i^\top\in\Real^{L}$ and
\[
\bA_i=\operatorname{TopR}(\mathbf{u}_i),\qquad |\bA_i|=r\ll L.
\]
We stack the selected directions
\[
\mathbf{U}_i=
\left[\bar{\zz}_{t,\ell_1}\cdots \bar{\zz}_{t,\ell_r}\right]^\top
\in\Real^{r\times d},\qquad \ell_k\in\bA_i,
\]
and obtain a thin QR factorization
\[
\mathbf{U}_i^\top=\QQ_i\mathbf{R}_i,\qquad
\QQ_i\in\Real^{d\times m},\quad \QQ_i^\top\QQ_i=I_m,\quad m\le r.
\]
The ACS projector is $\PPi_i^{\parallel}=\QQ_i\QQ_i^\top$.

As stated in the method, $\bA_i$ is a discrete index set and we do not backpropagate through $\operatorname{TopR}$. In our implementation we also treat $\QQ_i$ as constant w.r.t.\ $\zz_i$ during differentiation, i.e., $\frac{\partial \QQ_i}{\partial \zz_i}=0$, to avoid unstable gradients through QR under changing index sets. Therefore, gradients propagate only through subsequent matrix products, projections, normalizations, and dot-products.

\subsection{Projection, normalization, and subspace affinity}
\label{supp:diff_proj}

\mypar{Projected features.}
With $\PPi_i^{\parallel}$ fixed (given $\QQ_i$), projected query and gallery features are
\[
\zz_i^{\parallel}=\PPi_i^{\parallel}\zz_i,
\qquad
\zz_{g,j}^{\parallel}=\PPi_i^{\parallel}\zz_{g,j},
\]
and hence
\[
\frac{\partial \zz_i^{\parallel}}{\partial \zz_i}=\PPi_i^{\parallel}.
\]
We are able to treat $\PPi_i^{\parallel}$ as relatively constant with respect to $\zz_i$, due to the stop-gradient in the ACS basis construction.

\mypar{Stabilized normalization.}
We use $\tilde{\zz}=\frac{\zz}{\|\zz\|_2+\epsilon}$. For any nonzero $v\in\Real^d$, let $r(\vv)=\|\vv\|_2+\epsilon$ and $\tilde \vv=\vv/r(\vv)$. A convenient Jacobian form is
\[
\frac{\partial \tilde \vv}{\partial \vv}
=
\frac{1}{r(\vv)}
\left(
I - \frac{\vv \vv^\top}{\|\vv\|_2\,r(\vv)}
\right),
\]
which reduces to $\frac{1}{\|\vv\|_2}\left(I-\tilde \vv \tilde \vv^\top\right)$ when $\epsilon\to 0$ and $\|\vv\|_2>0$. Adding $\epsilon>0$ makes the denominator $r(\vv)$ bounded away from zero, so $\tilde \vv=\vv/r(\vv)$ does not explode when $\|\vv\|_2$ is very small (e.g., after projection). Consequently, the Jacobian terms that scale as $1/r(\vv)$ and $1/(\|\vv\|_2\,r(\vv))$ remain numerically controlled, preventing gradient blow-up near near-zero feature norms.

\mypar{Subspace affinity.}
The ACS subspace affinity is
\[
\bAA^{\parallel}[i,j] = (\tilde{\zz}_i^{\parallel})^\top \tilde{\zz}_{g,j}^{\parallel},
\qquad
\tilde{\zz}_i^{\parallel}=\operatorname{norm}(\zz_i^{\parallel}),\quad
\tilde{\zz}_{g,j}^{\parallel}=\operatorname{norm}(\zz_{g,j}^{\parallel}).
\]
Since $\ZZ_g$ is cached, $\tilde{\zz}_{g,j}^{\parallel}$ is constant w.r.t.\ $\theta$.
Let $v_i=\zz_i^{\parallel}$. Then
\[
\frac{\partial \bAA^{\parallel}[i,j]}{\partial v_i}
=
\left(\frac{\partial \tilde v_i}{\partial v_i}\right)^\top \tilde{\zz}_{g,j}^{\parallel},
\qquad
\frac{\partial \bAA^{\parallel}[i,j]}{\partial \zz_i}
=
\PPi_i^{\parallel}
\left(\frac{\partial \tilde v_i}{\partial v_i}\right)^\top \tilde{\zz}_{g,j}^{\parallel}.
\]

\subsection{Gradient of the final ACS logits}
\label{supp:diff_score}

The baseline identity affinity is
\[
\idmx[i,j] = (\tilde{\zz}_i)^\top \tilde{\zz}_{g,j},
\]
with $\tilde{\zz}_{g,j}$ constant and $\tilde{\zz}_i=\operatorname{norm}(\zz_i)$. Therefore,
\[
\frac{\partial \idmx[i,j]}{\partial \zz_i}
=
\left(\frac{\partial \tilde{\zz}_i}{\partial \zz_i}\right)^\top \tilde{\zz}_{g,j}.
\]

The final logits used by the loss are
\[
\bAA[i,j] = \idmx[i,j] + \alpha_{\mathrm{acs}}\,\bAA^{\parallel}[i,j].
\]
Thus
\[
\frac{\partial \bAA[i,j]}{\partial \zz_i}
=
\frac{\partial \idmx[i,j]}{\partial \zz_i}
+
\alpha_{\mathrm{acs}}\,\frac{\partial \bAA^{\parallel}[i,j]}{\partial \zz_i}.
\]

\subsection{End-to-end gradient to the adapted parameters summary}
\label{supp:diff_summary}

Let $\mathbf{g}_i=\frac{\partial \mathcal{L}_{\text{TTA}}}{\partial \mathbf{s}_i}\in\Real^{N_g}$ be the gradient w.r.t.\ logits for query $i$ (Sec.~\ref{supp:diff_loss}). Then
\[
\frac{\partial \mathcal{L}_{\text{TTA}}}{\partial \zz_i}
=
\sum_{j=1}^{N_g} g_{i,j}\,
\frac{\partial \bAA[i,j]}{\partial \zz_i}.
\]
Finally, since $\zz_i=\venc(x_i;\theta)$ for I2I-ReID (or $\zz_i=\tenc(\cdot;\theta_t)$ for T2I-ReID), we have established that gradients exist and are well defined for each step of \method{} which can be computed by autograd algorithm \cite{paszke2019pytorch}. Despite the stop-gradient and several top level indexing operations, the objective is differentiable almost everywhere and gradients propagate through: (1) matrix products for $\mathbf{u}_i$ (used only for selection), (2) linear projection $\zz\mapsto \PPi_i^{\parallel}\zz$, (3) stabilized normalization $\zz\mapsto \tilde{\zz}$, (4) inner products defining $\idmx$ and $\bAA^{\parallel}$, (5) softmax, and (piecewise constant) top-$k$ indexing in the loss.

\section{Full prompt bank constructions details}
\label{supp:sec:prompt_bank_construction}

\begin{figure}[t]
    \centering
    \includegraphics[width=0.98\columnwidth]{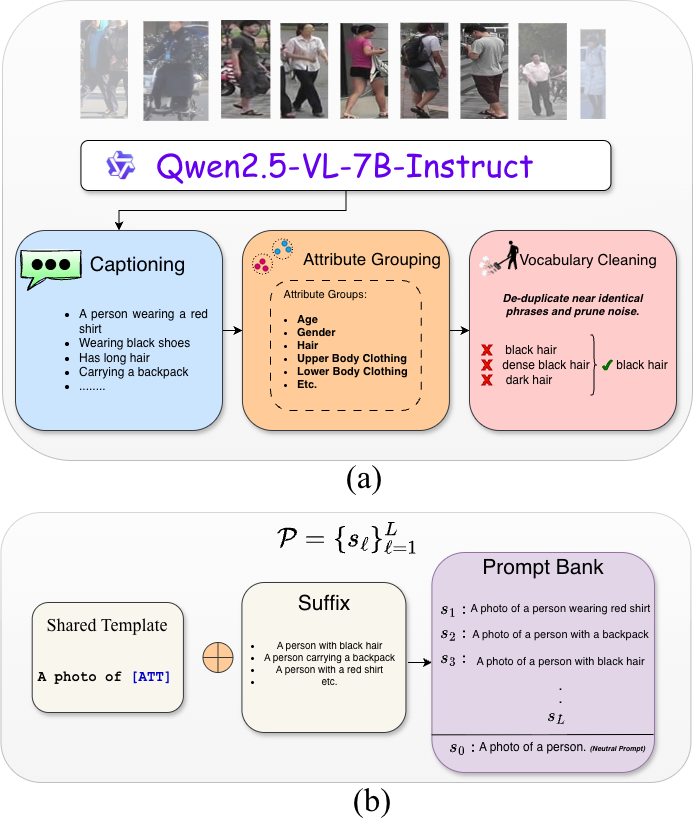}
    \caption{\textbf{Prompt-bank construction pipeline.} (a) A VLM is used to caption unlabeled pedestrian images and extract recurring appearance cues. 
    These cues are grouped into semantic categories and cleaned to remove duplicate or near-identical phrases. (b) The cleaned attribute suffixes are inserted into a shared template to form the final prompt bank $\mathcal{P}=\{s_\ell\}_{\ell=1}^{L}$, together with a neutral prompt $s_0$ used for prompt centering.}
    \label{fig:supp_prompt_bank_construction}
\end{figure}

This section describes the construction of the attribute prompt bank used by \method{}.
The overall pipeline is shown in \autoref{fig:supp_prompt_bank_construction}. Given a set of unlabeled pedestrian images, we first use a vision-language model to extract compact appearance descriptions, then group the extracted cues into predefined pedestrian attribute categories, clean duplicate or noisy phrases, and finally insert the resulting attribute suffixes into a shared prompt template. Importantly, the procedure does not use identity labels, attribute annotations for any kind of target-domain supervision. The resulting bank is a fixed transferable vocabulary of pedestrian appearance cues, which is encoded once and reused by \method{} during test-time adaptation.

\subsection{Input image collection and sampling}

Let $\mathcal{X}_{\mathrm{src}}=\{x_n\}_{n=1}^{N}$ denote a collection of pedestrian images used only for prompt-bank construction. In our default setting, $\mathcal{X}_{\mathrm{src}}$ is taken from the Market-1501 training split. For the prompt-bank source ablation, we repeat the same procedure using images from MSMT17, CUHK-PEDES, RSTP-ReID, and a manually specified prompt list. When person identifiers are available in the file names or dataset metadata, we sample at most a fixed number of images per identity to avoid over-representing identities with many camera views. These identifiers are used only for balanced image sampling and are never used as identity supervision for training or adaptation.

\subsection{VLM-based appearance extraction}

We use Qwen2.5-VL-7B-Instruct \cite{yang2025qwen3} to extract short, visually grounded pedestrian attributes from each sampled image. The instruction asks the model to ignore background, pose, action, camera information, brand names, and identity-specific information, and to return only compact appearance attributes. The output is constrained to a JSON-like schema so that the extracted cues can be parsed automatically and aggregated across the dataset.

\begin{tcolorbox}[
    colback=SkyBlue!6,
    colframe=SkyBlue!70!Black,
    title=\textbf{VLM appearance-extraction instruction},
    fonttitle=\bfseries,
    boxrule=0.7pt,
    arc=2pt,
    left=4pt,
    right=4pt,
    top=4pt,
    bottom=4pt
]
\small
\ttfamily
Extract only visible pedestrian appearance attributes from this cropped person image.\\[2pt]
Return ONLY valid JSON with these keys:\\
\{\\
\quad ``upper\_color'': ``string or list of short strings; use unknown if not visible'',\\
\quad ``upper\_type'': ``string or list of short strings; use unknown if not visible'',\\
\quad ``lower\_color'': ``string or list of short strings; use unknown if not visible'',\\
\quad ``lower\_type'': ``string or list of short strings; use unknown if not visible'',\\
\quad ``shoes'': ``string or list of short strings; use unknown if not visible'',\\
\quad ``bag'': ``string or list of short strings; use unknown if not visible'',\\
\quad ``accessory'': ``string or list of short strings; use unknown if not visible'',\\
\quad ``hair'': ``string or list of short strings; use unknown if not visible''\\
\}\\[2pt]
Rules: use lowercase; keep values short; do not mention background, pose, action, camera, brands, identity, or image quality; use ``unknown'' if a cue is absent or unclear; output JSON only.
\end{tcolorbox}

For analyses involving sensitive or demographic-like prompts, we additionally allow the keys \texttt{age} and \texttt{gender}. However, these categories are not required by the method. In the attribute-category ablation, we explicitly evaluate the effect of removing age and gender prompts to verify that \method{} does not depend on such cues.

\subsection{Attribute grouping}

The parsed VLM outputs are grouped into a small set of pedestrian-centric categories. For the default bank, we use appearance categories that are commonly visible in standard short-term ReID. When included for analysis, the additional categories are $\{\texttt{age},\texttt{gender}\}$. The purpose of grouping is not to predict supervised attributes, but to organize the mined vocabulary into interpretable categories from which query-conditioned ACS directions can be selected.

\begin{tcolorbox}[
    colback=Peach!8,
    colframe=Orange!80!Black,
    title=\textbf{Attribute grouping schema},
    fonttitle=\bfseries,
    boxrule=0.7pt,
    arc=2pt,
    left=4pt,
    right=4pt,
    top=4pt,
    bottom=4pt
]
\small
\begin{tabular}{ll}
\textbf{Category} & \textbf{Example extracted values} \\
\hline
\texttt{upper\_color} & red, black, white, blue, green \\
\texttt{upper\_type} & t-shirt, shirt, jacket, hoodie, coat \\
\texttt{lower\_color} & black, blue, gray, brown, white \\
\texttt{lower\_type} & pants, jeans, shorts, skirt, dress \\
\texttt{shoes} & white shoes, black shoes, sneakers \\
\texttt{bag} & backpack, handbag, shoulder bag \\
\texttt{accessory} & hat, glasses, umbrella \\
\texttt{hair} & short, long, black, brown \\
\texttt{age} & young, adult, elderly \\
\texttt{gender} & man, woman, person \\
\end{tabular}
\end{tcolorbox}

\subsection{Vocabulary cleaning and de-duplication}

The raw VLM outputs contain synonymous, near-duplicate, and occasionally noisy phrases. We therefore normalize each extracted value before adding it to the vocabulary. The cleaning stage lowercases all strings, removes punctuation and filler words, discards unknown or non-visible values, and merges common synonyms. For example, \texttt{grey} is canonicalized to \texttt{gray}, \texttt{trousers} to \texttt{pants}, and \texttt{back pack} to \texttt{backpack}. Long phrases are discarded because they usually indicate caption spillover rather than atomic attributes.

After normalization, we count the frequency of each attribute phrase within each category. Attributes that occur fewer than a threshold $m_{\min}$ are removed, and we optionally keep only the top $M$ phrases per category. This produces a compact bank that covers recurring pedestrian cues while suppressing rare or noisy descriptions. Formally, for category $c \in \mathcal{C}$, we retain
\begin{equation}
    \mathcal{A}_c
    =
    \left\{
    a \;:\; \mathrm{count}(a,c) \geq m_{\min}
    \right\},
\end{equation}
where $a$ is a cleaned attribute suffix.

\begin{tcolorbox}[
    colback=RedOrange!6,
    colframe=RedOrange!80!Black,
    title=\textbf{Vocabulary cleaning examples},
    fonttitle=\bfseries,
    boxrule=0.7pt,
    arc=2pt,
    left=4pt,
    right=4pt,
    top=4pt,
    bottom=4pt
]
\small
\resizebox{\columnwidth}{!}{%
\begin{tabular}{lll}
\textbf{Raw phrase} & \textbf{Cleaning operation} & \textbf{Canonical phrase} \\
\hline
\texttt{grey shirt} & spelling normalization & \texttt{gray shirt} \\
\texttt{back pack} & synonym merge & \texttt{backpack} \\
\texttt{trousers} & synonym merge & \texttt{pants} \\
\texttt{not visible} & unknown-value removal & discarded \\
\texttt{person wearing a red shirt} & filler removal & \texttt{red shirt} \\
\end{tabular}%
}
\end{tcolorbox}

\subsection{Prompt synthesis}

The cleaned attribute suffixes are converted into natural-language prompts using a shared template. For most categories, the generic form is
\begin{equation}
    s_\ell = \text{``A photo of a person with [ATT].''}
\end{equation}
where \texttt{[ATT]} is replaced by the cleaned attribute suffix. For grammatical consistency, we use category-specific templates for clothing colors, clothing types, shoes, bags, accessories, and hair. Examples are shown below.

\begin{tcolorbox}[
    colback=Lavender!10,
    colframe=Purple!75!Black,
    title=\textbf{Prompt synthesis templates},
    fonttitle=\bfseries,
    boxrule=0.7pt,
    arc=2pt,
    left=4pt,
    right=4pt,
    top=4pt,
    bottom=4pt
]
\small
\resizebox{\columnwidth}{!}{%
\begin{tabular}{ll}
\textbf{Category} & \textbf{Template} \\
\hline
\texttt{upper\_color} & \texttt{A photo of a person wearing a [ATT] upper-body garment.} \\
\texttt{upper\_type} & \texttt{A photo of a person wearing [ATT].} \\
\texttt{lower\_color} & \texttt{A photo of a person wearing a [ATT] lower-body garment.} \\
\texttt{lower\_type} & \texttt{A photo of a person wearing [ATT].} \\
\texttt{shoes} & \texttt{A photo of a person wearing [ATT] shoes.} \\
\texttt{bag} & \texttt{A photo of a person carrying [ATT].} \\
\texttt{accessory} & \texttt{A photo of a person with [ATT].} \\
\texttt{hair} & \texttt{A photo of a person with [ATT] hair.} \\
\end{tabular}}
\end{tcolorbox}

The final prompt bank is
\begin{equation}
    \mathcal{P}=\{s_\ell\}_{\ell=1}^{L},
\end{equation}
where each $s_\ell$ corresponds to one retained attribute phrase. We also define a neutral prompt
\begin{equation}
    s_0 = \text{``A photo of a person.''}
\end{equation}
During \method{}, all prompts are encoded by the frozen text encoder and centered by subtracting the neutral prompt embedding:
\begin{equation}
    \bar{\mathbf{z}}_{t,\ell}
    =
    \mathbf{z}_{t,\ell}
    -
    \mathbf{z}_{t,0}.
\end{equation}
This centering step reduces shared-template bias and makes the prompt bank behave as a set of attribute directions rather than a set of independent prompt classifiers.

\subsection{Prompt-bank construction and implementation details}
\label{sec:supp_prompt_bank_impl}

\paragraph{Attribute mining.}
We construct the attribute-prompt bank once as an offline preprocessing step using Qwen2.5-VL-7B-Instruct. For each ReID dataset, attributes are mined exclusively from its training-image partition; test queries and gallery images are never used for prompt-bank construction. For Market-1501, we sample at most four training images per identity and cap the total mining set at 3,000 images. Sampling is performed with a fixed random seed of $0$. Training-set identity identifiers are used only to obtain approximately balanced image sampling across identities; identity labels are never provided to the VLM, do not determine the extracted vocabulary, and are not used during PRAISE test-time adaptation. Unless otherwise stated, junk/background identities are excluded.

\paragraph{VLM inference.}
Images are processed independently with Qwen2.5-VL-7B-Instruct in evaluation mode using \texttt{bfloat16} precision on a single GPU. Generation is deterministic with a maximum of 192 newly generated tokens per image. The VLM is instructed to return only a JSON object containing short, visually observable pedestrian attributes and to output \texttt{unknown} when an attribute is not reliably visible. We explicitly prohibit descriptions of background, pose, action, camera, identity, brands, or image quality. By default, the mined vocabulary covers eight person-centric categories: upper-body color, upper-body clothing type, lower-body color, lower-body clothing type, shoes, bags, accessories, and hair. We do not include the optional age or gender categories in the experiments reported in this work.

\paragraph{Vocabulary cleaning and prompt generation.}
The raw VLM outputs are parsed and canonicalized before constructing the bank. Values marked as unknown/unclear are discarded, textual variants are normalized, common color and clothing synonyms are merged, and excessively long attribute phrases are removed. We retain an attribute only if it occurs in at least five mined images and retain at most 40 attributes per category. Duplicate prompts are subsequently removed. Each surviving attribute is converted into a category-specific natural-language template; for example, an upper-body color is expressed as ``A photo of a person wearing a \{attribute\} upper-body garment,'' while a bag attribute uses ``A photo of a person carrying \{attribute\}.'' The resulting bank contains $L=\textbf{57}$ unique prompts for Market-1501. The complete vocabulary and occurrence counts are provided in the released prompt-bank files.

\paragraph{Reproducibility and computational cost.}
For Market-1501, we sample up to four training images per identity, with a maximum of 3,000 images used for prompt mining. Attributes are retained only if they are observed in at least five sampled images, and the vocabulary is capped at 40 attributes per semantic category. All sampling is performed with a fixed random seed of 0. The complete mining procedure requires approximately \textbf{~102 minutes} (~12 minutes for loading model shards and ~90 minutes for analyzing 3000 images from the training dataset) on an \textbf{NVIDIA RTX A6000 GPU}. This cost is incurred only once when constructing the reusable semantic vocabulary and is therefore reported separately from online PRAISE inference. Once constructed, the prompt bank is stored and reused for all subsequent queries; prompt embeddings are likewise precomputed before test-time adaptation. Additional experiments in \autoref{supp:sec:source_abl} show that banks mined from a different source dataset remain effective, indicating that PRAISE does not rely on a dataset-specific prompt vocabulary.

\section{Additional Experiments}
\label{supp:sec:exp}

\subsection{Effectiveness of \method{} in addressing image corruption domain shifts}
\label{sec:supp:corrupt}

\begin{table}[!ht]
\centering
\caption{\textbf{Image Corruptions and Noise.} Comparison of TTA methods on \textbf{I2IReID} using a PLIP \cite{zuo2024plip} backbone. Results are reported on \textbf{Market-1501} under five corruptions (mAP / R1).}
\label{tab:tta_i2ireid_market_corrupt}
\setlength{\tabcolsep}{5pt}
\renewcommand{\arraystretch}{1.15}
\resizebox{\columnwidth}{!}{%
\begin{tabular}{l !{\vrule width 1.2pt} cc | cc | cc | cc | cc}
\toprule
\multirow{2}{*}{\textbf{TTA Method}}
& \multicolumn{10}{c}{\textbf{Market-1501 (Corrupted)}} \\
& \multicolumn{2}{c|}{\textbf{Gaussian Noise}}
& \multicolumn{2}{c|}{\textbf{Poisson Noise}}
& \multicolumn{2}{c|}{\textbf{Defocus Blur}}
& \multicolumn{2}{c|}{\textbf{Motion Blur}}
& \multicolumn{2}{c}{\textbf{Brightness}} \\
& \textbf{mAP} & \textbf{R1}
& \textbf{mAP} & \textbf{R1}
& \textbf{mAP} & \textbf{R1}
& \textbf{mAP} & \textbf{R1}
& \textbf{mAP} & \textbf{R1} \\
\midrule
No-adapt (Source) \venue{NeurIPS'24} & 24.2 & 50.7 & 14.5 & 37.5 & 13.7 & 35.7 & 8.9 & 26.1 & 33.5 & 60.9 \\
\midrule
TENT \cite{wang2020tent}   \venue{ICLR'21} & 23.4 & 48.0 & 9.0 & 22.6 & 11.5 & 34.2 & 6.9 & 19.9 & 29.7 & 55.3 \\
SHOT \cite{liang2020shot}  \venue{ICML'20} & 23.8 & 51.1 & 15.7 & 36.3 & 13.2 & 33.6 & 1.9 & 4.9 & 28.5 & 55.5 \\
SAR \cite{niu2023towards}  \venue{ICLR'23} & 26.6 & 52.2 & 16.8 & 38.8 & 14.2 & 35.0 & 8.7 & 24.3 & 31.8 & 58.1 \\
ETA \cite{niu2022efficient} \venue{ICML'22} & 25.1 & 50.8 & 16.4 & 39.1 & 14.0 & 34.8 & 9.3 & 26.1 & 31.1 & 57.7 \\
DeYO \cite{lee2024entropy}  \venue{ICLR'24} & 27.4 & 54.0 & 18.2 & 41.8 & 14.6 & 37.5 & 10.2 & 28.9 & 33.6 & 61.2 \\
\midrule
TPT \cite{shu2022test} \venue{NeurIPS'22} & 26.7 & 53.0 & 15.1 & 35.9 & 15.1 & 38.2 & 9.1 & 23.6 & 34.1 & 61.2 \\
TDA \cite{karmanov2024efficient} \venue{CVPR'24} & 26.7 & 53.8 & 16.2 & 39.4 & 13.7 & 35.9 & 8.9 & 26.1 & 33.4 & 60.8 \\
\midrule
TEMP \cite{adachi2024test} \venue{IJCNN'24} & 25.1 & 51.2 & 15.1 & 35.7 & 14.1 & 35.9 & 9.1 & 26.2 & 33.6 & 61.1 \\
Anchor-ReID \cite{almansoori2023anchor} \venue{SCIA'23} & 11.0 & 24.8 & 5.9 & 19.2 & 8.5 & 16.8 & 5.5 & 13.4 & 22.2 & 32.3 \\
Norm \cite{song2025exploring} \venue{ICLR'25} & 24.8 & 50.9 & 14.6 & 37.8 & 14.5 & 36.1 & 9.4 & 26.8 & 35.1 & 62.9 \\
\midrule
\colrow
\method{} (Ours) & 27.7 & 55.1 & 15.3 & 38.9 & 23.9 & 41.8 & 15.7 & 33.4 & 44.8 & 69.8 \\
\bottomrule
\end{tabular}}
\vspace*{-6pt}
\end{table}

\autoref{tab:tta_i2ireid_market_corrupt} evaluates I2I-ReID on Market-1501 under five common corruptions, taking insights from common TTA evaluations in \cite{wang2020tent, shu2022test, lafon2025cliptta}, originally proposed in \cite{hendrycks2019benchmarking}. Two regimes emerge: \textbf{(i) noise corruptions} (Gaussian/Poisson) where several generic entropy-based TTA methods become comparatively stronger than in the clean-domain table, and \textbf{(ii) blur/brightness corruptions} where relative rankings more closely match the main-paper trends. The application of these corruptions can be summarized as follows:

\begin{itemize}
    \item \textbf{Gaussian noise.} We convert the image to a tensor in $[0,1]$, we add i.i.d.\ Gaussian noise $\mathcal{N}(0,\sigma^2)$ per pixel/channel with $\sigma=0.06$, and we clamp back to $[0,1]$.

    \item \textbf{Poisson noise.} We convert the image to a tensor $x\in[0,1]$, we scale intensities to pseudo-counts $x\cdot \lambda$ with $\lambda=20$, we sample elementwise Poisson noise and rescale by $1/\lambda$, and we clamp to $[0,1]$.

    \item \textbf{Defocus blur.} We apply PIL Gaussian blur with radius $r=3.0$.

    \item \textbf{Motion blur.} We build a normalized $k{\times}k$ line kernel with $k=13$ at a randomly chosen angle in $\{0,45,90,135\}$, we reflect-pad the image, and we convolve per-channel using grouped convolution, then clamp to $[0,1]$.

    \item \textbf{Brightness.} We apply a brightness transform with factor $f\sim\mathcal{U}(0.55,1.45)$ using PIL brightness enhancement.
\end{itemize}

Under additive noise, the dominant failure mode is often \emph{feature dispersion} (reduced similarity gaps) rather than semantic misalignment. Entropy-based adaptation methods  can then act as effective \emph{confidence restorers}: by sharpening the retrieval posterior they partially recover top-neighbor separation, which explains their improved mAP/R1 under Gaussian and Poisson noise compared to their clean-domain behavior. This is especially visible for SAR, ETA and DeYO, which surpass or match the no-adapt baseline under both noise types (e.g., DeYO reaches $27.4/54.0$ on Gaussian and $18.2/41.8$ on Poisson, while the baseline is $24.2/50.7$ and $14.5/37.5$). In other words, when corruptions are mostly \emph{non-semantic} and approximately independent and identically distributed, entropy filtering tends to select ``still-recoverable'' queries, and sharpening helps re-separate nearest neighbors.

Some methods exhibit extreme behavior under specific corruptions (e.g., SHOT collapses under Motion Blur with $1.9/4.9$). This is plausible because motion blur introduces strong, structured high-frequency attenuation and spatial streaking, which can systematically break pseudo-label/consistency assumptions and thus updates can become dominated by incorrect neighbors, leading to rapid confirmation-bias collapse. Conversely, methods can appear unusually strong under noise because the corruption perturbs features in a way that preserves coarse identity structure and the TTA objective can then effectively ``denoise'' the retrieval posterior by amplifying pre-existing similarity gaps.

For Defocus Blur, Motion Blur, and Brightness, the ordering becomes more consistent with the main paper: generic entropy-minimization baselines provide only modest gains (or can degrade), while retrieval-aligned objectives are more reliable. Notably, \method{} improves substantially on the harder structured corruptions (e.g., Defocus Blur $23.9/41.8$ and Motion Blur $15.7/33.4$), where naive sharpening is more likely to overfit to spurious top-1 matches. This is consistent with the view that blur/illumination changes induce more \emph{semantics-affecting} shifts (texture suppression, edge degradation, global intensity shifts) that demand retrieval-specific inductive biases, making the robustness trends closer to those observed in the evaluations of the main manuscript.

\subsection{Attribute bank Source and Category Ablations}\label{supp:sec:source_abl}

\begin{table}[t]
\centering
\caption{\textbf{Effect of attribute prompt-bank source.} We evaluate \method{} using the same fixed hyperparameters while varying only the source used to construct the attribute prompt bank. The manual bank is hand-designed, while the remaining banks are mined from the corresponding ReID datasets using the same label-free captioning and cleaning protocol. Results show that \method{} is not tied to a specific prompt-bank source and remains stable across transferable pedestrian attribute vocabularies.
}
\label{tab:prompt_bank_source_ablation}
\setlength{\tabcolsep}{4.5pt}
\renewcommand{\arraystretch}{1.08}
\resizebox{\columnwidth}{!}{
\begin{tabular}{l|cccc|cccc}
\toprule
\multirow{2}{*}{\textbf{Prompt-bank source}}
&
\multicolumn{2}{c}{\textbf{MSMT17}}
&
\multicolumn{2}{c}{\textbf{Market-1501}}
&
\multicolumn{2}{c}{\textbf{CUHK-PEDES}}
&
\multicolumn{2}{c}{\textbf{RSTP-ReID}}
\\
\cmidrule(lr){2-3}
\cmidrule(lr){4-5}
\cmidrule(lr){6-7}
\cmidrule(lr){8-9}
&
\textbf{mAP}
&
\textbf{R1}
&
\textbf{mAP}
&
\textbf{R1}
&
\textbf{mAP}
&
\textbf{R1}
&
\textbf{mAP}
&
\textbf{R1}
\\
\midrule
No-adapt source model
& 10.5 & 34.2
& 35.0 & 62.6
& 46.4 & 55.9
& 23.1 & 34.5
\\
\midrule
\rowcolor{Gray!10}
Generic prompt bank
& 17.8 & 36.3
& 45.9 & 70.1
& 48.7 & 56.1
& 27.1 & 36.1
\\
\rowcolor{SkyBlue!10}
Market-1501 mined bank
& 18.2 & 38.5
& 46.5 & 71.1
& 49.6 & 58.4
& 27.6 & 37.6
\\
\rowcolor{SkyBlue!10}
MSMT17 mined bank
& 18.4 & 39.1
& 46.5 & 70.1
& 49.2 & 57.6
& 26.9 & 37.0
\\
\rowcolor{SeaGreen!10}
CUHK-PEDES mined bank
& 17.9 & 36.5
& 45.4 & 69.7
& 49.4 & 58.1
& 27.6 & 37.5
\\
\rowcolor{SeaGreen!10}
RSTP-ReID mined bank
& 17.8 & 36.3
& 45.3 & 69.7
& 49.1 & 57.7
& 27.2 & 36.9
\\
\bottomrule
\end{tabular}
}
\end{table}

\noindent \textbf{Impact of attribute source.} \autoref{tab:prompt_bank_source_ablation} studies whether the performance of \method{} depends on the dataset used to construct the attribute prompt bank. Across all four benchmarks, replacing the default Market-1501 mined bank with manually designed prompts or banks mined from MSMT17, CUHK-PEDES, and RSTP-ReID leads to only small performance changes. All prompt banks consistently improve over the no-adapt source model, indicating that the gains of \method{} do not come from a dataset-specific vocabulary or from overfitting the prompt bank to a particular benchmark. We observe a slight advantage for prompt banks mined from I2I-ReID datasets, especially Market-1501 and MSMT17. This is likely because these datasets contain surveillance-style pedestrian crops collected across multiple cameras, viewpoints, poses, and illumination conditions, which closely matches the visual gallery distribution used during ReID retrieval. As a result, the mined attributes tend to capture common visually grounded pedestrian cues, such as clothing color, clothing type, bags, shoes, and hair, that transfer well across both I2I and T2I settings. In contrast, banks mined from T2I-ReID datasets remain competitive but can reflect the visual and linguistic biases of text-person-search datasets, where some cues may be more description-oriented or less diverse in surveillance appearance. Nevertheless, the small variation across rows shows that \method{} is robust to the prompt-bank source and only requires a transferable pedestrian attribute vocabulary rather than a dataset-specific prompt set.

\begin{table}[t]
\centering
\caption{\textbf{Attribute-category ablation.} We evaluate \method{} using the same Market-1501 mined prompt bank and fixed hyperparameters, while removing one group of attributes at a time. This isolates which attribute categories contribute most to the query-conditioned ACS geometry.
}
\label{tab:attribute_category_ablation}
\setlength{\tabcolsep}{5.5pt}
\renewcommand{\arraystretch}{1.08}
\resizebox{0.8\columnwidth}{!}{
\begin{tabular}{lcccc}
\toprule
\multirow{2}{*}{\textbf{Prompt-bank variant}}
&
\multicolumn{2}{c}{\textbf{Market-1501}}
&
\multicolumn{2}{c}{\textbf{RSTP-ReID}}
\\
\cmidrule(lr){2-3}
\cmidrule(lr){4-5}
&
\textbf{mAP}
&
\textbf{R1}
&
\textbf{mAP}
&
\textbf{R1}
\\
\midrule
No-adapt source model
& 35.0 & 62.6
& 23.1 & 34.5
\\
\midrule
Full prompt bank
& 46.5 & 71.1
& 27.6 & 37.6
\\
\rowcolor{Gray!8}
w/o upper-body color
& 45.8 & 68.1
& 26.1 & 35.2
\\
\rowcolor{Gray!8}
w/o lower-body color
& 45.9 & 69.3
& 27.0 & 36.9
\\
\rowcolor{Gray!8}
w/o upper-body clothing
& 46.0 & 69.5
& 27.2 & 37.4
\\
\rowcolor{Gray!8}
w/o lower-body clothing
& 46.3 & 70.1
& 27.5 & 37.5
\\
\rowcolor{SeaGreen!8}
w/o bag/accessories
& 45.5 & 68.2
& 25.5 & 34.3
\\
\rowcolor{SeaGreen!8}
w/o hair
& 46.5 & 70.3
& 27.7 & 37.8
\\
\rowcolor{RedOrange!8}
w/o age
& 46.7 & 70.8
& 27.6 & 37.6
\\
\rowcolor{RedOrange!8}
w/o gender
& 46.4 & 70.2
& 27.5 & 37.6
\\
\bottomrule
\end{tabular}
}
\end{table}

\noindent \textbf{Impact of attribute categories.} \autoref{tab:attribute_category_ablation} studies the contribution of different attribute groups by removing one category at a time from the Market-1501 mined prompt bank. All ablated variants remain well above the no-adapt source model, showing that \method{} does not depend on a single attribute type and instead benefits from a redundant set of complementary pedestrian cues. Among the categories, upper-body color and bag/accessory prompts have the largest effect: removing upper-body color reduces performance on both Market-1501 and RSTP-ReID, while removing bag/accessory prompts produces one of the strongest drops, especially on RSTP-ReID. This is consistent with standard ReID behavior, where upper clothing color is often the most visible short-term identity cue, and carried objects or accessories provide discriminative evidence when present. In contrast, removing hair has negligible effect, suggesting that hair cues are either less reliably visible or already captured indirectly by other appearance prompts. Importantly, removing age or gender does not degrade performance and in some cases slightly improves it, indicating that \method{} does not rely on demographic-like attributes. This supports the use of a safer prompt bank focused primarily on clothing, color, and carried-object cues in applications surrounding sensitive private information.

\subsection{Extended Performance on CLIP-ReID\cite{li2023clip} and RDE \cite{qin2024noisy}}

\mypar{Evaluation protocol.} The absolute no-adapt scores in \autoref{tab:tta_i2ireid_clipreid_source_training} and \autoref{tab:tta_t2ireid_source_training} should not be interpreted as reproduced numbers from the original papers. Our goal is to evaluate test-time adaptation under a common retrieval interface that is compatible with \method{}. Specifically, \method{} requires query, gallery, and text-prompt representations to lie in the same VLM embedding space so that prompt-induced directions can be used to construct ACS. We therefore use the CLIP image embedding with cosine similarity for retrieval, without the task-specific ReID head used in the original supervised evaluation. This changes the absolute no-adapt performance relative to the original paper, but it is applied uniformly to all methods in the table. Thus, the comparison measures how each TTA method modifies the same fixed-gallery cosine retrieval geometry, rather than comparing against separately optimized task-specific heads.

\begin{table}[t]
\centering
\caption{Comparison of TTA methods on \textbf{I2I-ReID} benchmarks using a source-trained CLIP-ReID \cite{li2023clip} image encoder. All methods are evaluated under the same PRAISE-compatible cosine-retrieval protocol, with fixed gallery embeddings and no target identity labels.}
\vspace*{-5pt}
\label{tab:tta_i2ireid_clipreid_source_training}
\setlength{\tabcolsep}{3.2pt}
\renewcommand{\arraystretch}{1.12}
\resizebox{\columnwidth}{!}{%
\begin{tabular}{l !{\vrule width 1.1pt} cccc | cccc}
\toprule
\multirow{4}{*}{\textbf{TTA Method}}
&
\multicolumn{8}{c}{\textbf{I2I-ReID}}
\\
\cmidrule(lr){2-9}
&
\multicolumn{4}{c|}{\textbf{Train: MSMT17}}
&
\multicolumn{4}{c}{\textbf{Train: Market-1501}}
\\
\cmidrule(lr){2-5}
\cmidrule(lr){6-9}
&
\multicolumn{2}{c}{\textbf{Test: MSMT17}}
&
\multicolumn{2}{c|}{\textbf{Test: Market-1501}}
&
\multicolumn{2}{c}{\textbf{Test: MSMT17}}
&
\multicolumn{2}{c}{\textbf{Test: Market-1501}}
\\
&
\textbf{mAP} & \textbf{R1}
&
\textbf{mAP} & \textbf{R1}
&
\textbf{mAP} & \textbf{R1}
&
\textbf{mAP} & \textbf{R1}
\\
\midrule
No-adapt (Source)
& \ul{57.4} & \ul{79.9} & 37.8 & 63.3
& \ul{21.0} & \ul{46.9} & 83.6 & 92.6
\\
\midrule
TENT \cite{wang2020tent} \venue{ICLR'21}
& 55.2 & 75.4 & 35.2 & 60.5
& 18.6 & 44.2 & 82.1 & 91.1
\\
SHOT \cite{liang2020shot} \venue{ICML'20}
& 49.7 & 68.4 & 32.1 & 57.6
& 14.6 & 41.2 & 80.6 & 89.8
\\
SAR \cite{niu2023towards} \venue{ICLR'23}
& 55.3 & 76.8 & 37.6 & 63.2
& 19.7 & 45.2 & 83.1 & 92.4
\\
ETA \cite{niu2022efficient} \venue{ICML'22}
& 55.2 & 76.8 & 37.1 & 62.8
& 20.1 & 45.8 & 82.7 & 92.2
\\
DeYO \cite{lee2024entropy} \venue{ICLR'24}
& 55.3 & 76.9 & 37.9 & 63.3
& 19.8 & 41.3 & 83.8 & 92.7
\\
\midrule
TPT \cite{shu2022test} \venue{NeurIPS'22}
& 57.1 & 79.5 & 38.1 & 63.9
& 20.2 & 46.1 & 83.9 & 92.8
\\
TDA \cite{karmanov2024efficient} \venue{CVPR'24}
& 55.7 & 78.8 & 37.9 & 63.5
& 20.1 & 45.8 & 83.6 & 92.7
\\
\midrule
TEMP \cite{adachi2024test} \venue{IJCNN'24}
& 55.1 & 75.3 & 38.9 & \ul{65.1}
& 18.9 & 44.7 & 84.2 & 93.1
\\
Norm \cite{song2025exploring} \venue{ICLR'25}
& 57.2 & 79.5 & \ul{41.1} & \bf 65.8
& 20.7 & 46.4 & \ul{85.0} & \ul{93.4}
\\
\midrule
\colrow
\method{} (Ours)
& \bf 63.1 & \bf 83.8 & \bf 42.4 & \ul{65.1}
& \bf 24.3 & \bf 50.9 & \bf 86.3 & \bf 94.3
\\
\bottomrule
\end{tabular}}
\vspace*{-6pt}
\end{table}

\mypar{How does \method{} perform on CLIP-ReID \cite{li2023clip} backbone.} \autoref{tab:tta_i2ireid_clipreid_source_training} evaluates \method{} in a more strongly supervised setting using a source-trained CLIP-ReID backbone. Since the backbone has already been trained for identity discrimination, the source model provides a much stronger retrieval geometry than the generic PLIP setting, leaving less room for test-time adaptation. Under this setting, several generic TTA objectives degrade performance, especially TENT, SHOT, SAR, ETA, and DeYO. This behavior is expected in open-set ReID: entropy-style sharpening can reinforce the model's current nearest-neighbor structure, which may already be highly confident but still biased by camera, pose, or dataset-specific appearance correlations. As a result, directly sharpening the retrieval distribution can over-concentrate probability mass around visually similar distractors rather than improving identity ranking. In contrast, \method{} consistently improves the source-trained CLIP-ReID model across both source and cross-dataset evaluations. The gains are moderate compared to the generic-backbone setting, but this is expected because the supervised backbone already produces strong identity features. Importantly, \method{} improves both source-domain performance and cross-dataset transfer, indicating that the ACS-guided adaptation does not simply compensate for a weak backbone. Instead, the attribute-conditioned subspace provides complementary semantic structure that can refine an already discriminative ReID embedding. Norm remains a strong post-hoc baseline and achieves the best Rank-1 in one cross-dataset case, but \method{} obtains the best mAP in all settings and the best or second-best Rank-1 across the table. These results show that \method{} remains useful even when applied on top of a competitive source-trained ReID backbone, addressing the concern that the improvements are specific to weaker or generic VLM representations.

\begin{table}[t]
\centering
\caption{Comparison of TTA methods on \textbf{T2I-ReID} benchmarks using source-trained RDE \cite{qin2024noisy} backbones. All methods are evaluated under the same PRAISE-compatible cosine-retrieval protocol, with fixed gallery embeddings and no target identity labels.}
\label{tab:tta_t2ireid_source_training}

\vspace*{-5pt}
\setlength{\tabcolsep}{3.4pt}
\renewcommand{\arraystretch}{1.12}
\resizebox{\columnwidth}{!}{%
\begin{tabular}{l !{\vrule width 1.1pt} cccc | cccc}
\toprule
\multirow{4}{*}{\textbf{TTA Method}}
&
\multicolumn{8}{c}{\textbf{T2I-ReID}}
\\
\cmidrule(lr){2-9}
&
\multicolumn{4}{c|}{\textbf{Train: CUHK-PEDES}}
&
\multicolumn{4}{c}{\textbf{Train: RSTP-ReID}}
\\
\cmidrule(lr){2-5}
\cmidrule(lr){6-9}
&
\multicolumn{2}{c}{\textbf{Test: CUHK}}
&
\multicolumn{2}{c|}{\textbf{Test: RSTP}}
&
\multicolumn{2}{c}{\textbf{Test: CUHK}}
&
\multicolumn{2}{c}{\textbf{Test: RSTP}}
\\
&
\textbf{mAP} & \textbf{R1}
&
\textbf{mAP} & \textbf{R1}
&
\textbf{mAP} & \textbf{R1}
&
\textbf{mAP} & \textbf{R1}
\\
\midrule
No-adapt (Source) \venue{NeurIPS'24}
& 64.2 & 72.7 & 41.2 & 54.9
& 33.5 & 36.8 & 47.9 & 62.1
\\
\midrule
TENT \cite{wang2020tent} \venue{ICLR'21}
& 64.0 & 72.8 & 41.0 & 53.4
& 33.1 & 36.4 & 47.5 & 62.0
\\
SHOT \cite{liang2020shot} \venue{ICML'20}
& 64.5 & 72.9 & 41.3 & 54.7
& 33.8 & 37.1 & 48.3 & 62.5
\\
\midrule
TPT \cite{shu2022test} \venue{NeurIPS'22}
& 64.2 & 72.9 & 41.4 & 55.1
& 33.6 & 36.9 & 48.1 & 62.5
\\
TDA \cite{karmanov2024efficient} \venue{CVPR'24}
& 59.1 & 68.7 & 38.4 & 51.2
& 31.2 & 35.7 & 46.1 & 61.1
\\
WATT\,(S) \cite{osowiechi2024watt} \venue{NeurIPS'24}
& 59.8 & 69.2 & 39.1 & 51.7
& 31.3 & 35.8 & 46.9 & 61.4
\\
WATT\,(P) \cite{osowiechi2024watt} \venue{NeurIPS'24}
& 61.2 & 70.2 & 39.9 & 52.2
& 31.6 & 36.0 & 46.4 & 61.8
\\
CLIPArTT \cite{hakim2024clipartt} \venue{WACV'25}
& 59.6 & 69.1 & 38.7 & 51.1
& 31.1 & 35.5 & 46.2 & 60.9
\\
CLIPTTA \cite{lafon2025cliptta} \venue{NeurIPS'25}
& 62.6 & 71.2 & 40.2 & 52.7
& 31.4 & 35.9 & 46.5 & 61.8
\\
\midrule
TCR \cite{li2024test} \venue{ICLR'25}
& \ul{64.6} & \ul{73.1} & \ul{42.5} & \ul{55.7}
& \ul{34.1} & \ul{37.3} & \ul{48.5} & \ul{63.1}
\\
\midrule
\colrow
\method{} (Ours)
& \bf 67.1 & \bf 75.0 & \bf 46.3 & \bf 60.9
& \bf 38.1 & \bf 42.8 & \bf 50.6 & \bf 65.5
\\
\bottomrule
\end{tabular}}

\vspace*{-4pt}
\end{table}

\mypar{How does \method{} perform on RDE \cite{qin2024noisy} backbone.} \autoref{tab:tta_t2ireid_source_training} evaluates \method{} on source-trained RDE backbones under both source-domain and cross-dataset T2I-ReID settings. Compared with the PLIP setting in the main paper, the no-adapt RDE model already provides a much stronger retrieval representation, so the available room for test-time adaptation is smaller. This is reflected in the relatively mild gains of most generic TTA methods, and in several cases their performance degradation. In particular, classification-oriented or generic VLM adaptation methods are not necessarily aligned with the fixed-gallery ranking objective of T2I-ReID, and their sharpening or prompt-adaptation objectives can disrupt an already discriminative retrieval embedding.

Among prior methods, \cite{li2024test} is the strongest baseline and achieves the second-best result across all evaluated metrics. Nevertheless, \method{} consistently improves over both the no-adapt RDE model and all TTA baselines. The gains are moderate on source-domain evaluation, where the supervised RDE representation is already well calibrated, and larger under cross-dataset transfer, where query-gallery alignment is more affected by dataset shift. This shows that ACS-guided adaptation is not limited to weak or generic VLM backbones; it remains complementary to a source-trained T2I-ReID model by refining the retrieval geometry using query-conditioned attribute directions.

\subsection{Gallery-size scalability}
\label{supp:sec:gallery_scaling}

\begin{figure}[t]
    \centering
    \includegraphics[width=0.8\linewidth]{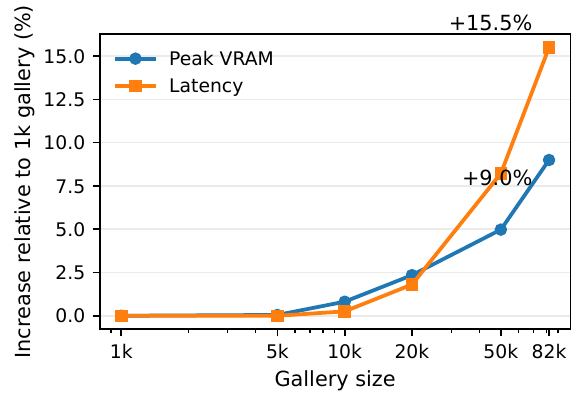}
    \caption{\textbf{Gallery-size scalability of \method{} on MSMT17 with PLIP.} Peak GPU-memory and adaptation-latency increases are reported relative to the 1k-gallery setting. The gallery is increased from 1,000 to the full 82,161 images while retaining the default adaptation batch size of 32. Even at the full gallery, peak memory increases by only $9.0\%$, while amortized per-query latency increases by $15.5\%$.}
    \label{fig:gallery_scaling}
\end{figure}

To evaluate the scalability of \method{} with respect to gallery size, we conduct an additional experiment on MSMT17 using the PLIP backbone. MSMT17 is particularly suitable for this analysis because its large gallery allows us to evaluate \method{} over a substantially wider range of gallery sizes than the other benchmarks. We progressively increase the gallery from 1,000 images to the full 82,161 images while keeping the adaptation configuration fixed, including the default query batch size of 32.

\autoref{fig:gallery_scaling} reports the relative increase in peak GPU memory and adaptation latency with respect to the 1k-gallery setting. We normalize both quantities to their respective values at 1,000 gallery images to make their scaling behavior directly comparable despite having different physical units. Peak GPU memory increases only modestly as the gallery grows: from 4,179\,MB at 1k images to 4,555\,MB for the complete 82,161-image gallery, corresponding to an increase of approximately $9.0\%$. Runtime exhibits similarly gradual scaling. The measured adaptation throughput decreases from 3.95 to 3.42 batches/s between the 1k and full-gallery settings. Since each adaptation iteration processes a batch of 32 queries, this corresponds to an increase in amortized latency from approximately 7.91 to 9.14\,ms/query, or $15.5\%$.

These results indicate that the gallery-dependent operations in \method{} introduce relatively small additional memory and computational overhead even when the gallery size is increased by more than $80\times$. Importantly, these measurements correspond only to online test-time adaptation and retrieval; one-time preprocessing operations such as gallery feature extraction and prompt-bank construction are reported separately and are not included in this scalability analysis.

\subsection{Sensitivity to Initial Retrieval Quality}
\label{supp:sec:initial_rank}

A potential failure mode of query-side adaptation is confirmation bias, where an incorrect initial neighborhood is reinforced during optimization. To study this effect, we stratify Market-1501 queries according to the initial rank of the first valid cross-camera match under the frozen PLIP baseline. We consider six groups: Rank-1, ranks 2--5, 6--10, 11--20, 21--50, and $51+$. For each group, we record whether \method{} improves, preserves, or degrades the rank of the first correct match. We also report the change in reciprocal rank,
\[
\Delta\mathrm{RR}=\frac{1}{r_{\mathrm{\text{\method{}}}}}-\frac{1}{r_{\mathrm{\text{base}}}},
\]
which emphasizes changes near the top of the retrieval list while reducing the influence of very large absolute rank changes.

\begin{figure}[t]
    \centering
    \includegraphics[width=\linewidth]{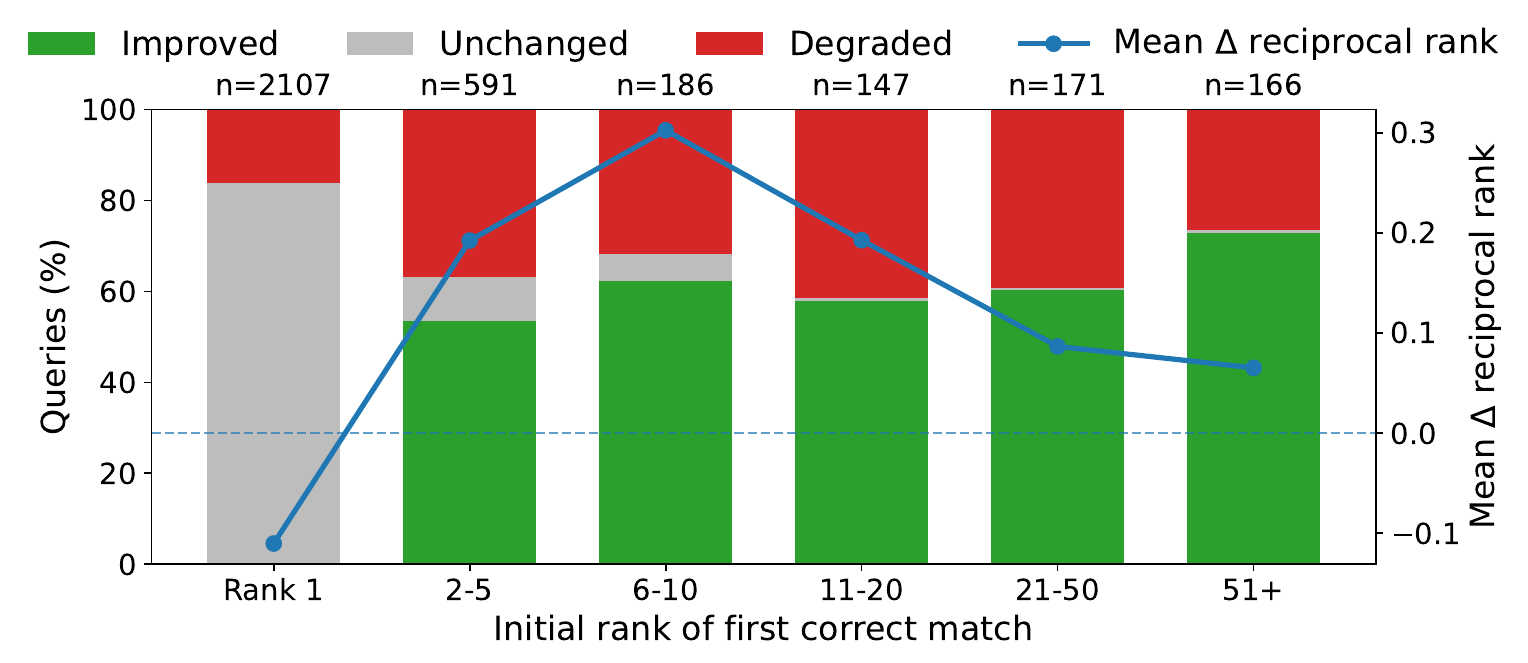}
    \caption{\textbf{Sensitivity to initial retrieval quality on Market-1501.} Queries are grouped by the initial rank of the first valid cross-camera match under the frozen PLIP baseline. Bars show the percentage of queries whose first-correct rank improves, remains unchanged, or degrades after PRAISE. The line reports mean $\Delta\mathrm{RR}$, where positive values indicate improved retrieval.}
    \label{fig:supp_initial_rank}
\end{figure}

As shown in \autoref{fig:supp_initial_rank}, \method{} is not restricted to queries whose correct identity already appears in the initial retrieval head. Excluding queries that are already correct at Rank-1, a majority of queries improve in every initial-rank group. When the first correct match lies outside the initial top-5 neighborhood, \method{} improves its rank for \textbf{63.4\%} of queries. Among these cases, \textbf{30.2\%} are moved into the final top-5 and \textbf{13.3\%} reach Rank-1. The median rank gain is six positions and the mean reciprocal-rank change is \textbf{+0.164}. Positive mean $\Delta\mathrm{RR}$ is observed across all non-Rank-1 groups.

For queries already correct at Rank-1, \method{} preserves Rank-1 in \textbf{83.9\%} of cases. Since these queries cannot improve further in first-match rank, any change can only preserve or degrade their position. Overall, these results show that \method{} does not simply reinforce candidates already present in the initial top-$k$ neighborhood and can recover a substantial fraction of initially poor retrievals. Degradation remains possible for unreliable neighborhoods, which we retain as a limitation of query-side test-time adaptation.

\subsection{Sensitivity to Direct Fusion Weight}
\label{supp:sec:fusion_weight}

To assess whether the weakness of direct attribute-score fusion is caused by the choice of fusion weight, we sweep $\lambda_{\mathrm{att}}$ in Eq.~(2) over $\{0,0.1,0.2,0.5,1.0\}$. Here, $\lambda_{\mathrm{att}}=0$ corresponds to identity-only adaptation without attribute-score fusion. As shown in \autoref{tab:supp_fusion_weight}, a small attribute contribution improves mAP, with $\lambda_{\mathrm{att}}=0.1$ reaching 39.2 on Market-1501 and 24.5 on RSTPReID. However, Rank-1 accuracy decreases from 65.0 to 60.3 and from 35.4 to 32.9, respectively. Increasing the fusion weight further degrades both metrics. Thus, direct attribute scores contain useful retrieval information, but no tested fusion weight consistently improves both mAP and Rank-1 over the identity-only setting. All settings also remain well below PRAISE, supporting the use of attributes as query-conditioned geometric guidance rather than as directly added retrieval scores.

\begin{table}[t]
\centering
\caption{\textbf{Sensitivity to the direct attribute-fusion weight $\lambda_{\mathrm{att}}$.} Results are reported as mAP/R1.}
\label{tab:supp_fusion_weight}
\resizebox{0.55\linewidth}{!}{
\begin{tabular}{c|cc}
\toprule
$\lambda_{\mathrm{att}}$ & Market-1501 & RSTPReID \\
\midrule
0.0 & 36.8/65.0 & 23.5/35.4 \\
0.1 & 39.2/60.3 & 24.5/32.9 \\
0.2 & 38.9/60.1 & 24.3/32.5 \\
0.5 & 37.5/58.7 & 23.7/30.8 \\
1.0 & 35.1/54.5 & 23.1/28.3 \\
\bottomrule
\end{tabular}}
\end{table}

\subsection{Impact on camera bias}

\begin{figure}[t]
  \centering
  \includegraphics[width=0.8\linewidth]{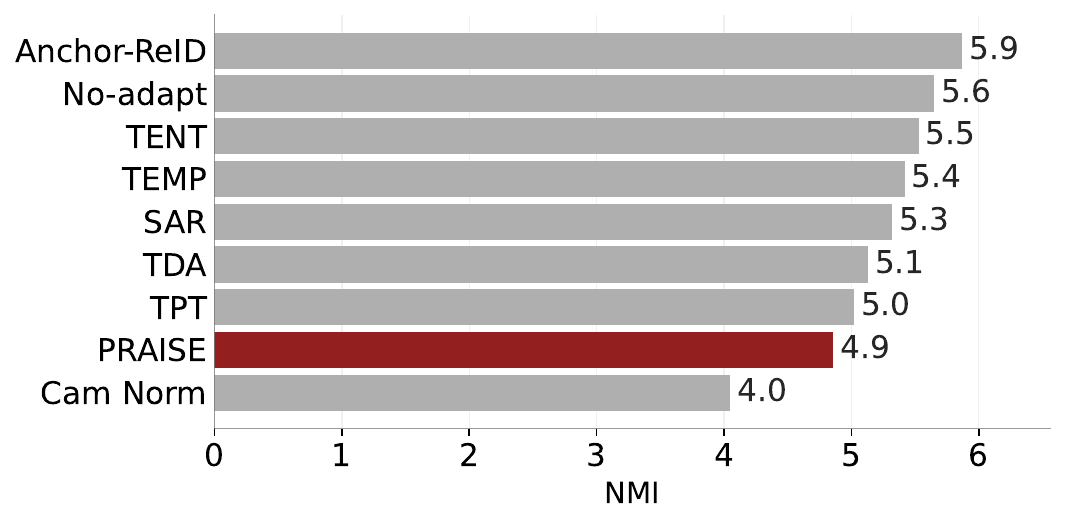}
  \vspace{-6pt}
  \caption{NMI comparison across methods.}
  \label{fig:nmi_barplot}
\end{figure}

We want to assess whether our approach effectively mitigates camera bias. To measure this, we employ a mutual information (MI) based evaluation following the protocol suggested in \cite{song2025exploring}. The rationale behind this experiment is that, if feature representations exhibit reduced dependence on camera labels, the mutual information between features and camera identifiers should correspondingly decrease. \autoref{fig:nmi_barplot} presents the outcomes of this analysis. The results indicate that both Camera Normalization \cite{song2025exploring} and \method{} achieve the lowest normalized mutual information (NMI) scores, suggesting a comparatively weaker association between features and camera labels. Camera Normalization proves to be more effective when it comes to dealing with camera bias. While these findings do not constitute definitive evidence for the complete elimination of camera bias, they provide strong empirical support that our approach substantially attenuates such bias in the learned feature representations.

\subsection{Visualization: Failure Cases}
\label{supp:vis_fail}

\begin{figure*}[!t]
    \centering
    \includegraphics[width=0.6\linewidth]{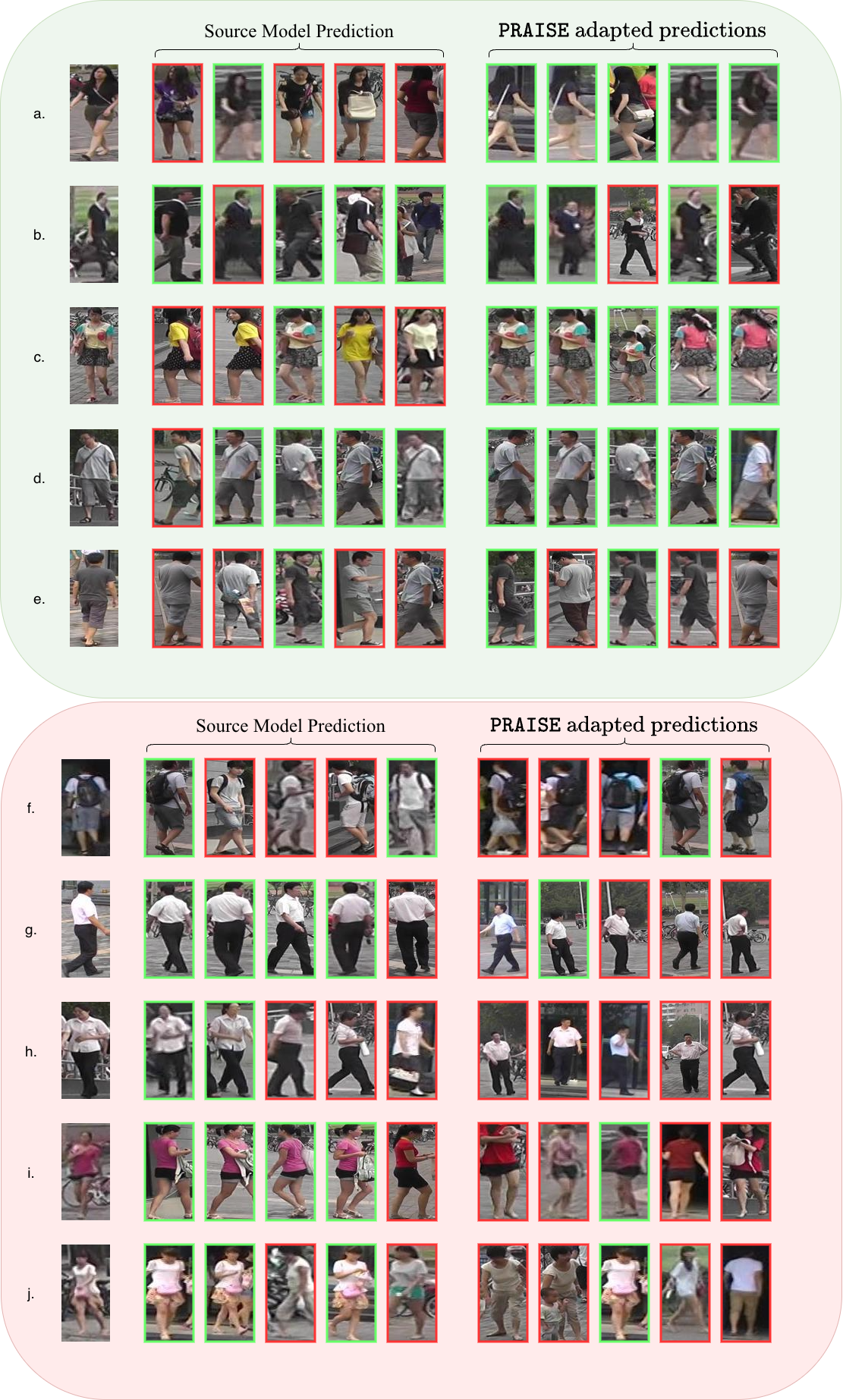}
    \caption{Some success and failure cases for I2I-ReID from Market-1501 \cite{zheng2015scalable} dataset.}
    \label{fig:fail_i2i}
\end{figure*}

\begin{figure*}[!t]
    \centering
    \includegraphics[width=0.6\linewidth]{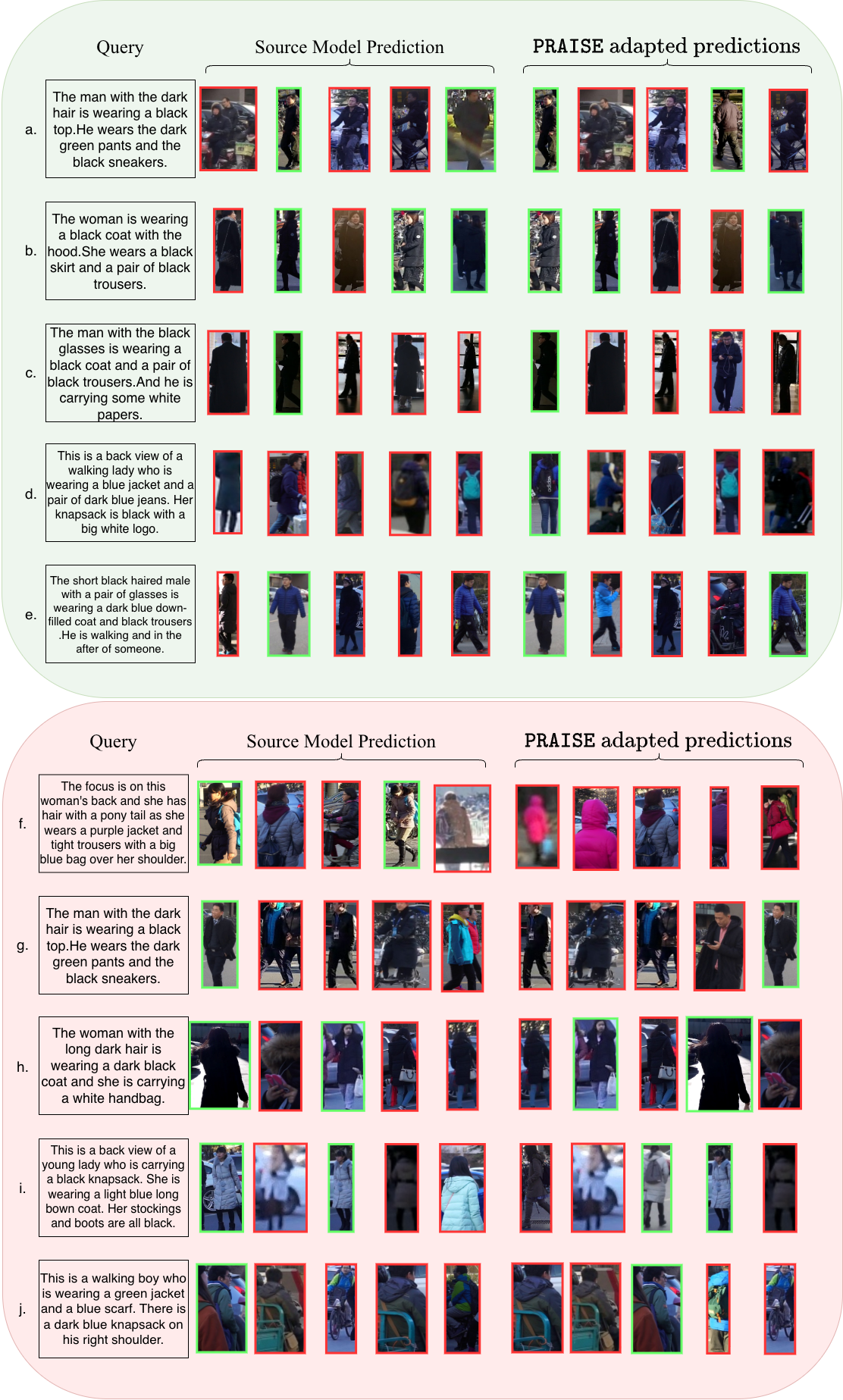}
    \caption{Some success and failure cases for T2I-ReID from RSTPReID \cite{zhu2021dssl} dataset.}
    \label{fig:fail_t2i}
\end{figure*}

\autoref{fig:fail_i2i} presents qualitative I2I-ReID examples on Market-1501. In the success cases among the first five rows, the adapted rankings better prioritize the true identity by suppressing visually similar distractors. This is visible when the source model retrieves near-duplicates with mismatched clothing patterns or accessories, while the adapted predictions shift more correct matches into the top positions.

The failure cases suggest a characteristic failure mode of online adaptation with attribute guidance. When the initial attribute-driven affinity is incorrect, the subsequent sharpening can reinforce the wrong neighborhood and bias the update toward an incorrect set of gallery candidates. In (h) and (j), even though the query depicts a woman, the adapted retrieval settles on men with similar color cues. This is consistent with an erroneous early gender-related attribute estimate that redirects the ACS subspace toward spurious cues. Overall, the attribute directions are often beneficial, but they can occasionally sway retrieval by fixating on odd or weakly discriminative cues, leading to self-reinforced mistakes after adaptation.

\autoref{fig:fail_t2i} shows analogous behavior for T2I-ReID on RSTPReid. Some challenging queries remain correct after adaptation. In (c) and (d), the descriptions are visually ambiguous and still the adapted predictions maintain or improve the correct matches, indicating that ACS can help focus similarity on text-consistent visual evidence rather than incidental distractors. Failure cases for T2I arise when adaptation aggravates noisy correspondence between the text and the image set. In (f) and (i), the adapted rankings drift further from the correct identity, which is consistent with sharpening amplifying an initially misaligned top set. Compared to I2I, we observe that gender-related errors are better contained in T2I, likely because explicit textual descriptors provide a stronger constraint that reduces the chance of flipping into an incorrect gender-driven neighborhood.

\section{Additional Implementation Details}
\label{sec:impl_concise}

\subsection{Sharpness-Aware and Reliable Adaptation (SAR)}
\label{sec:sar_short}
\begin{itemize}[leftmargin=*,itemsep=2pt,topsep=2pt]
    \item \textbf{Idea.} Apply entropy-minimization TTA only on \emph{reliable} queries; use a SAM-style sharpness-aware update to avoid brittle minima.
    \item \textbf{ReID posterior.} Treat \emph{gallery indices} as transient pseudo-classes: $\PP=\operatorname{softmax}(\idmx/\tau_{\text{id}})$ and $H_i=-\sum_j \PP[i,j]\log \PP[i,j]$.
    \item \textbf{Efficiency.} Compute entropy on a top-$K$ candidate set per query: $\mathcal{N}_K(i)=\operatorname{TopK}(\idmx[i,:])$, renormalize on $\mathcal{N}_K(i)$.
    \item \textbf{Filtering.} Update only if $H_i<\gamma$; skip step if no samples pass.
    \item \textbf{Optimization.} Adapt only normalization affine parameters in $\venc$; SAM perturbation $\epsilon(\theta)=\rho\,g/(\|g\|_2+\varepsilon)$ with two-pass update.
    \item \textbf{Stability.} Episodic resets per batch (or EMA-based recovery in non-episodic variants).
\end{itemize}

\subsection{Efficient Test-Time Adaptation (ETA/EATA)}
\label{sec:eta_short}
\begin{itemize}[leftmargin=*,itemsep=2pt,topsep=2pt]
    \item \textbf{Idea.} Entropy-based TTA with two filters: (i) keep confident queries; (ii) drop \emph{redundant} queries whose posteriors look like a running mean.
    \item \textbf{Posterior/entropy.} Same retrieval posterior as SAR (optionally top-$K$ renormalized) to compute $H_i$.
    \item \textbf{Filter 1 (reliability).} Keep $\mathcal{I}_1=\{i: H_i<\gamma\}$.
    \item \textbf{Filter 2 (redundancy).} Maintain running mean posterior $\bar{\pp}$; keep $\mathcal{I}_2=\{i\in\mathcal{I}_1: |\cos(\PP[i,:],\bar{\pp})|<\delta\}$ (implemented on a shared candidate support when using top-$K$).
    \item \textbf{Loss.} Reweighted entropy: $w_i=\exp(-(H_i-\gamma))$, $\mathcal{L}_{\text{ETA}}=\frac{1}{\sum_{i\in\mathcal{I}_2}w_i}\sum_{i\in\mathcal{I}_2}w_iH_i$.
    \item \textbf{What is tuned.} Only normalization affine parameters in $\venc$ (episodic reset recommended).
\end{itemize}

\subsection{DeYO for ReID (Entropy + PLPD filtering)}
\label{sec:deyo_short}
\begin{itemize}[leftmargin=*,itemsep=2pt,topsep=2pt]
    \item \textbf{Idea.} Keep confident queries (entropy filter) and further keep those whose top prediction is \emph{sensitive} to a mild corruption (PLPD), which aims to discard unreliable gradients.
    \item \textbf{Posterior.} $\PP=\operatorname{softmax}(\idmx/\tau_{\text{id}})$ (often computed on top-$K$ indices for speed).
    \item \textbf{Entropy filter.} $\mathcal{I}_{\text{ent}}=\{i: H_i<\gamma_{\text{Ent}}\}$.
    \item \textbf{PLPD.} Apply corruption $\mathcal{T}$, compute $\PP'$, take $j_i^\star=\arg\max_j\PP[i,j]$, and $\mathrm{PLPD}_i=\PP[i,j_i^\star]-\PP'[i,j_i^\star]$.
    \item \textbf{PLPD filter.} Keep $\mathcal{I}_{\text{plpd}}=\{i\in\mathcal{I}_{\text{ent}}: \mathrm{PLPD}_i>\gamma_{\text{PLPD}}\}$.
    \item \textbf{Loss.} Reweighted entropy on $\mathcal{I}_{\text{plpd}}$ with weights combining entropy-margin and PLPD terms.
    \item \textbf{What is tuned.} Only normalization affine parameters in $\venc$; use mild $\mathcal{T}$ to avoid uniform-collapse of $\PP'$.
\end{itemize}

\subsection{Test-time Prompt Tuning (TPT) for ReID (I2I and T2I)}
\label{sec:tpt_short}
\begin{itemize}[leftmargin=*,itemsep=2pt,topsep=2pt]
    \item \textbf{Idea.} Adapt an \emph{attribute prompt bank} (not the encoders) by sharpening prompt posteriors across stochastic views, then use prompt-induced attribute similarity as an auxiliary retrieval signal.
    \item \textbf{Prompts.} Precompute $\ZZ_t^{(0)}=\tenc(\tokn(\mathcal{S}))$ and optionally center with neutral $s_0$; cache centered/normalized bank.
    \item \textbf{What is tuned.} Prompt-bank adapter $\Delta(\phi)$ (often low-rank), producing $\ZZ_t(\phi)=\operatorname{Norm}(\widetilde{\ZZ}_t^{(0)}+\Delta(\phi))$.
    \item \textbf{TPT loss.} For $M$ views (image augs for I2I; caption templates/perturbations for T2I), form $\qq^{(m)}=\operatorname{softmax}(\ZZ^{(m)}\ZZ_t(\phi)^\top/\tau_t)$ and minimize $H(\bar{\qq})$ with $\bar{\qq}=\frac{1}{M}\sum_m\qq^{(m)}$ (optionally keep low-entropy views only).
    \item \textbf{Retrieval.} Compute $\idmx=\ZZ_q\ZZ_g^\top$ and attribute similarity: 
    
    $\attmx=(\operatorname{softmax}(\ZZ_q\ZZ_t(\phi)^\top/\tau_t))(\operatorname{softmax}(\ZZ_g\ZZ_t(\phi)^\top/\tau_t))^\top$; fuse $\bAA=\idmx+\beta\attmx$.
    \item \textbf{Stability/efficiency.} Episodic reset of $\phi$; chunked $\ZZ_g\ZZ_t(\phi)^\top$ for large $N_g$.
\end{itemize}

\subsection{Training-free Dynamic Adapter (TDA) for T2I ReID}
\label{sec:tda_short}
\begin{itemize}[leftmargin=*,itemsep=2pt,topsep=2pt]
    \item \textbf{Idea.} No gradients: maintain \emph{positive/negative caches} of confident retrieval outcomes and correct logits by similarity-weighted scatter add/subtract.
    \item \textbf{Base logits.} $\mathbf{s}=(qG^\top)/\tau_{\text{id}}$ for text query features $q$ and cached gallery $G$.
    \item \textbf{Gating.} Use (top-$k$) retrieval entropy to decide whether a query is reliable enough to write to the positive cache, or uncertain enough to write distractors to the negative cache.
    \item \textbf{Logit correction.} Add weighted votes to cached positive indices and suppress cached negative candidate sets using $A(\mathrm{sim})=\alpha\exp(-\beta(1-\mathrm{sim}))$.
    \item \textbf{Practical.} Episodic clears caches per batch; non-episodic accumulates over stream; gate thresholds dominate behavior.
\end{itemize}

\subsection{TCR for T2I ReID (Query-shift adaptation with constraints)}
\label{sec:tcr_short}
\begin{itemize}[leftmargin=*,itemsep=2pt,topsep=2pt]
    \item \textbf{Idea.} Adapt the \emph{text encoder} (query side) using retrieval-derived pseudo-positives and constraints that enforce source-like cross-modal gap and text uniformity.
    \item \textbf{What is tuned.} Only LayerNorm affine parameters in $\tenc$ (image encoder and gallery are fixed).
    \item \textbf{Pseudo-pairs.} Retrieve top gallery item per caption and gather corresponding image features to form an in-batch $B\times B$ ``refined'' similarity.
    \item \textbf{Reliability.} Use entropy of the $B\times B$ refined posterior as confidence (low entropy = stable pseudo-positive).
    \item \textbf{Constraints.} Maintain a small queue of source-like items to estimate an adaptive entropy threshold and a target modality-gap statistic; optimize a weighted sum of noise-robust entropy, text-uniformity, and gap-matching terms.
\end{itemize}

\subsection{WATT for T2I ReID (Weight-averaged TTA)}
\label{sec:watt_short}
\begin{itemize}[leftmargin=*,itemsep=2pt,topsep=2pt]
    \item \textbf{Idea.} Run short TTA for multiple caption-template ``views'' and \emph{average} the resulting adapted normalization parameters (and optionally ensemble embeddings) to reduce template sensitivity.
    \item \textbf{What is tuned.} Only normalization affine parameters in $\tenc$; force dropout to eval mode.
    \item \textbf{Objective.} Define a posterior over gallery indices via $p=\operatorname{softmax}(z_qG^\top/\tau_{\text{id}})$ (or ACS variant) and optimize an unsupervised sharpening/soft-target loss.
    \item \textbf{Averaging.} Snapshot parameters per template and average (parallel) or apply sequentially; evaluate using the final averaged state.
\end{itemize}

\subsection{CLIPTTA-style TTA for T2I ReID (Contrastive with pseudo-positives)}
\label{sec:cliptta_short}
\begin{itemize}[leftmargin=*,itemsep=2pt,topsep=2pt]
    \item \textbf{Idea.} Build pseudo-positive image prototypes by retrieval and optimize a CLIP-style in-batch contrastive objective on caption--pseudoimage pairs.
    \item \textbf{What is tuned.} Only normalization affine parameters in $\tenc$ (gallery and image encoder fixed).
    \item \textbf{Pseudo-positives.} From $S=z_qG^\top$, select top-1 or weighted top-$k$ gallery features per caption (stop-grad through selection).
    \item \textbf{Loss.} Contrastive loss on the resulting $B\times B$ logits; fallback entropy-based objectives when $B=1$ and optional transductive soft-target variants for stability.
    \item \textbf{Optional memory.} FIFO queue of past pseudo-positives can be re-encoded under current $\tenc$ to stabilize updates across batches.
\end{itemize}
\end{document}